\documentclass[sn-chicago]{sn-jnl}

\usepackage{graphicx}%
\usepackage{multirow}%
\usepackage{amsmath,amssymb,amsfonts}%
\usepackage{amsthm}%
\usepackage{mathrsfs}%
\usepackage{xcolor}%
\usepackage{manyfoot}%
\usepackage{arydshln}
\usepackage{amsmath}
\DeclareMathOperator*{\argmax}{arg\,max}

\usepackage{hyperref}
\usepackage[normalem]{ulem}
\usepackage{booktabs}
\usepackage{subfig}
\usepackage{tabularray}

\makeatletter
\newcommand\uuuline{\bgroup\markoverwith%
   {%
     \textcolor{black}{\rule[-0.5ex]{2pt}{0.4pt}}%
     \llap{\textcolor{red}{\rule[-0.9ex]{2pt}{0.4pt}}}%
     \llap{\textcolor{darkgreen}{\rule[-1.3ex]{2pt}{0.4pt}}}%
   }%
   \ULon}
 \makeatother

 \makeatletter
\newcommand{\rAB}[2]{\bgroup \UL@setULdepth
 \markoverwith{\lower\ULdepth\hbox
   {\kern-.03em\vbox{\color{#1}\hrule width.2em\kern1.2\p@\color{#2}\hrule}\kern-.03em}}%
 \ULon}
  \makeatother
   
\definecolor{darkblue}{rgb}{0, 0, 0.5}
\definecolor{darkgreen}{rgb}{0, 0.5, 0}
\hypersetup{colorlinks=true,citecolor=darkblue, linkcolor=darkblue, urlcolor=darkblue}

\usepackage{relsize}

\newcommand{\CODE}[0]{{\smaller {\sf CODE}}}
\newcommand{\DATETIME}[0]{{\smaller {\sf DATETIME}}}
\newcommand{\DEM}[0]{{\smaller {\sf DEM}}}
\newcommand{\LOC}[0]{{\smaller {\sf LOC}}}
\newcommand{\MISC}[0]{{\smaller {\sf MISC}}}
\newcommand{\ORG}[0]{{\smaller {\sf ORG}}}
\newcommand{\PERSON}[0]{{\smaller {\sf PERSON}}}
\newcommand{\QUANTITY}[0]{{\smaller {\sf QUANTITY}}}
\newcommand{\ENT}[1]{{\smaller {\sf #1}}}

\newtheorem{example}{Example}%

\begin{document}

\title[Article Title]{Neural Text Sanitization with Privacy Risk Indicators: An Empirical Analysis}


\author*[1]{\fnm{Anthi} \sur{Papadopoulou}}\email{anthip@ifi.uio.no}

\author[2]{\fnm{Pierre} \sur{Lison}}\email{plison@nr.no}

\author[2]{\fnm{Mark} \sur{Anderson}}\email{anderson@nr.no}

\author[1]{\fnm{Lilja} \sur{\O{}vrelid}}\email{liljao@ifi.uio.no}

\author[2]{\fnm{Ildik\'{o}} \sur{Pil\'{a}n}}\email{pilan@nr.no}

\affil*[1]{\orgdiv{Language Technology Group}, \orgname{University of Oslo}, \orgaddress{\street{Gaustadalléen 23B}, \postcode{0373} \city{Oslo}, \country{Norway}}}

\affil[2]{\orgdiv{Norwegian Computing Center}, \orgaddress{\street{Gaustadalléen 23A}, \postcode{0373} \city{Oslo}, \country{Norway}}}


\abstract{Text sanitization is the task of redacting a document to mask all occurrences of (direct or indirect) personal identifiers, with the goal of concealing the identity of the individual(s) referred in it. In this paper, we consider a two-step approach to text sanitization and provide a detailed analysis of its empirical performance on two recently published datasets: the Text Anonymization Benchmark \citep{benchmark} and a collection of Wikipedia biographies \citep{bootstrapping}. The text sanitization process starts with a privacy-oriented entity recognizer that seeks to determine the text spans expressing identifiable personal information. This privacy-oriented entity recognizer is trained by combining a standard named entity recognition model with a gazetteer populated by person-related terms extracted from Wikidata. The second step of the text sanitization process consists in assessing the privacy risk associated with each detected text span, either isolated or in combination with other text spans. We present five distinct indicators of the re-identification risk, respectively based on language model probabilities, text span classification, sequence labelling, perturbations, and web search. We provide a contrastive analysis of each privacy indicator and highlight their benefits and limitations, notably in relation to the available labeled data.}




\maketitle

\section{Introduction}
\label{sec:intro}

The volume of text data available online is continuously increasing and constitutes an essential resource for the development of large language models (LLM). This trend has important privacy implications. Most text documents do, indeed, contain \textit{personally identifiable information} (PII) in one form or another -- that is, information that can directly or indirectly lead to the identification of a human individual. PII can be split in two main categories \citep{soton399692}:
\begin{description}
    \item \textbf{Direct identifiers}, which are information that can directly and irrevocably identify an individual. This includes for example a person's name, passport number, email address, phone number, username or home address. 
    \item \textbf{Quasi identifiers}, also called \textit{indirect} identifiers, which are not \textit{per se} sufficient to single out an individual, but may do so when combined with other quasi-identifiers. Examples of quasi-identifiers include the person's nationality, occupation, gender, place of work or date of birth. A well-known illustration of the privacy risks posed by quasi-identifiers comes from \citet{golle2006} who showed that between 63\% to 78\% of the US population could be uniquely identified by merely combining their gender, date of birth and zip code. 
\end{description}

Although some PII may be benign (if the text conveys e.g.~trivial or public information), it is far from always being the case. Documents such as court judgments, medical records or interactions with social services are often highly sensitive, and their disclosure may have catastrophic consequences for the individuals involved. Editing those documents to conceal the identity of those individuals is thus often desirable from both an ethical and legal perspective. This is precisely the goal of text sanitization \citep{chakaravarthy2008efficient,Sanchez2016,lison-etal-2021-anonymisation}.

\begin{figure}[t]
\centering
\includegraphics[width = 0.95\textwidth]{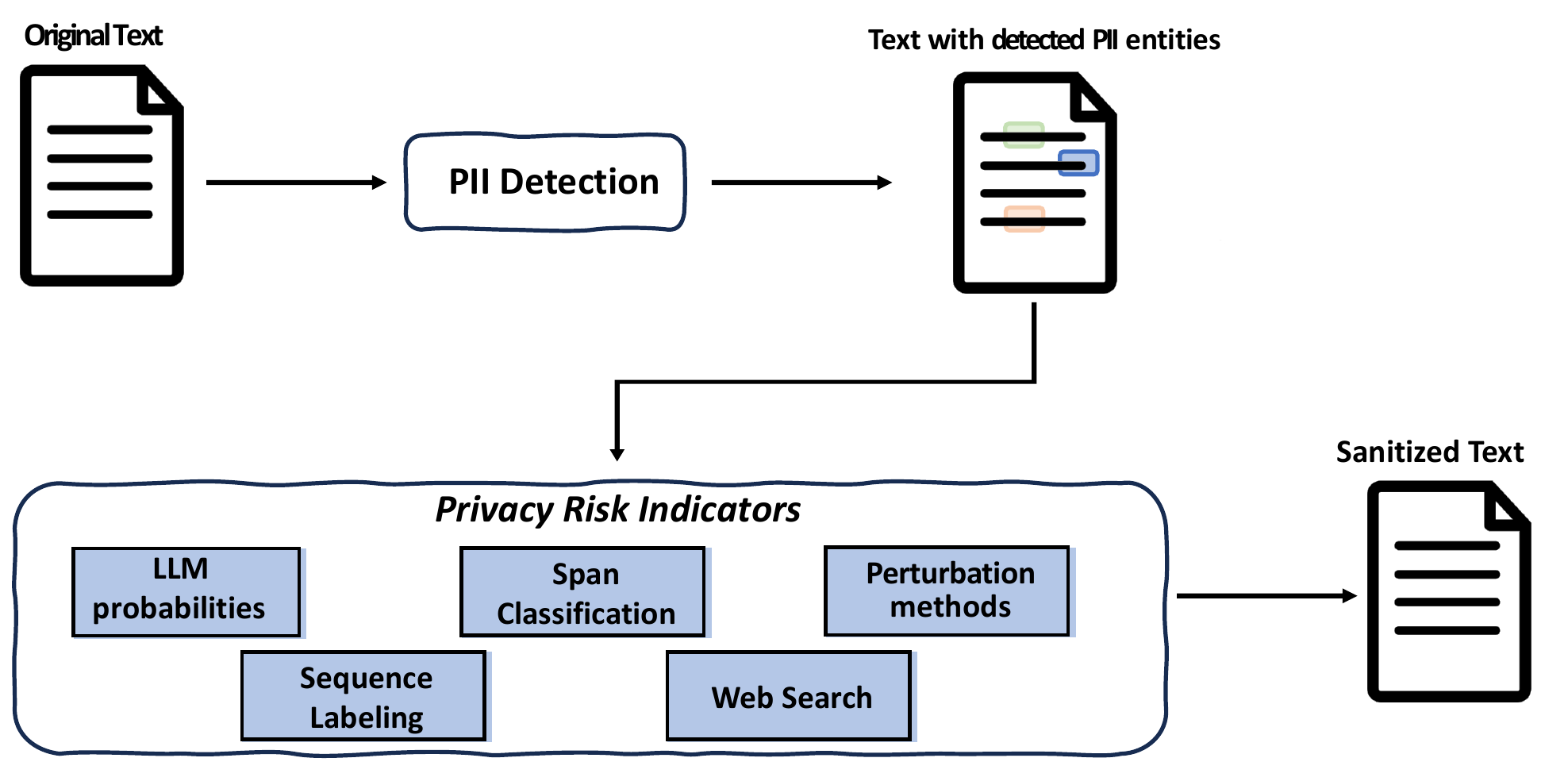}
\vspace{2mm}
\caption{Overview of the approaches presented and discussed in this paper. For a text document or a collection of text documents to sanitize, we first detect text spans expressing PII in the text. This can be done using a privacy-oriented entity recognizer. A set of five distinct privacy risk indicators can then be applied to assess whether the detected spans need to be masked. Finally, the text spans predicted to constitute a high re-identification risk are masked from the document.}
\label{fig:overview}
\end{figure}

In this paper, we present a two-step approach to the problem of text sanitization and provide a detailed analysis of its performance, based on two recently released datasets \citep{benchmark,bootstrapping}. The approach first seeks to detect text spans that convey personal information, using a neural sequence labelling model that combines named entity recognition (NER) with a gazetteer. The gazetteer is built by inferring from Wikidata a set of properties typically used to characterize human individuals, such as ``\textit{position held}" or ``\textit{manner of death}", and traversing the knowledge graph to extract all possible values for those properties. In the second step, the privacy risk of the detected text spans is assessed according to various indicators. Finally, the text spans that are deemed to constitute an unacceptable risk according to those indicators are masked. An overview of the approach is illustrated in Figure \ref{fig:overview}.

The paper makes the following four contributions:
\begin{enumerate}
    \item A privacy-oriented entity recognizer making it possible to detect PII information beyond named entities, using large lists of person-related terms extracted from Wikidata, extending previous work by \citet{papadopoulou-etal-2022-neural}.
    \item A quantitative and qualitative evaluation of the performance of this privacy-oriented entity recognizer for various types of PII.
    \item The development of five privacy risk indicators, based on LLM probabilities, span classification, perturbations, sequence labelling and web search.
    \item A comparative analysis of those privacy risk indicators, both in isolation and in combination, on the two aforementioned datasets.
\end{enumerate}

The paper consists of the following sections. Section \ref{sec:bg} provides a general background on data privacy and text sanitization. 
We then describe in Section \ref{sec:data} the two annotated datasets for text sanitization employed for the empirical analysis of this paper. In Section \ref{sec:er} we present, evaluate and discuss a privacy-oriented entity recognizer for the task of detecting PII in text documents. We present the set of five privacy risk indicators in Section \ref{sec:indicators}. The empirical performance of those indicators on the two datasets is then analysed in Section \ref{sec:eval}. We conclude in Section \ref{sec:conclusion}.

\section{Background}
\label{sec:bg}

\subsection{Definitions}

The right to privacy is a fundamental human right, as evidenced by its inclusion in the Universal Declaration of Human Rights and the European Convention on Human Rights. In the digital sphere, data privacy is enforced through multiple national and international regulations, such as the General Data Protection Regulation (GDPR) in Europe, the California Consumer Privacy Act (CCPA) in the United States or China's Personal Information Protection Law (PIPL). Although those regulations differ in both scope and interpretation, their common principle is that individuals should remain in control of their own data. In particular, the processing of personal data must have a legal ground, and cannot be shared to third parties without the explicit and informed consent of the person(s) the data refers to. 

One alternative strategy is to \textit{anonymize} data to ensure the data is no longer personal, and therefore out of the scope of privacy regulations. Anonymization, according to the GDPR, refers to the complete and irrevocable removal of all information that may directly or indirectly lead to re-identification. However, as shown by \citet{weitzenboeck2022gdpr}, transforming the data to make it completely anonymous is almost impossible to achieve in practice for unstructured data such as text, unless the content of the text is radically altered or the original source of the document is deleted. 

Although complete anonymization is hard to attain, text sanitization is a crucial tool to adhere with the general requirement of \textit{data minimization} which is enshrined in GDPR and most privacy regulations \citep{goldsteen2021data}.  The principle of data minimization states that one should only collect and retain the personal data that is strictly necessary to fulfill a given purpose.

The process of editing text documents to conceal the identity of a person has a somewhat confusing terminology \citep{lison-etal-2021-anonymisation,benchmark}. The GDPR makes use of the term \textit{pseudonymization} to denote a process of transforming data to conceal at least some personal identifiers, but in a way that does not amount to complete anonymization. The term \textit{de-identification} is also common \citep{article,johnson2020deidentification}, especially for work on medical patient records. De-identification approaches are typically restricted to the recognition of predefined entities, such as the categories of \citet{HIPAA}. In contrast, we define \textit{text sanitization} as the process of detecting and masking \textit{any} type of personal information in a text document that can lead to identification of the individual whose identity we wish to protect. 

Text sanitization is a topic of investigation in several research fields, notably in natural language processing (NLP) and in privacy-preserving data publishing (PPDP). Approaches to text rewriting based on differential privacy have also been proposed. We review below those approaches. 

\subsection{NLP Approaches}
\label{ssec:nlp}

NLP approaches to text sanitization have mainly focused on sequence labelling approaches, inspired by the large body of work on Named Entity Recognition. Such approaches aim at the detection of text spans containing personal identifiers \citep{Chi:Nic:16,Lam:Bal:Sub:16}. Most research works in this field to date have focused on the medical domain, where the Health Insurance Portability and Accountability Act of 1996 \citep{HIPAA} offers concrete rules that allow for the standardization of this task. HIPAA defines a set of Protected Health Information (PHI) data types that encompass direct identifiers (such as names or social security numbers) as well as  domain-specific demographic attributes including treatments received and health conditions. A wide variety of NLP methods have been developed for this task, including rule-based, machine learning-based and hybrid approaches \citep{sweeney1996replacing,neamatullah2008automated,YANG2015S30,Yog:May:Pfa:2018}.
Character-based recurrent neural networks \citep{dernoncourt2017identification,liu2017identification} and transformer architectures have also been investigated for this purpose \citep{johnson2020deidentification}. A recent initiative focused on  replacing sensitive information is INCOGNITUS \citep{ribeiro-etal-2023-incognitus}, a clinical note de-identification tool. The system allows for redacting documents with either a NER-based method or with an embedding-based approach substituting all tokens with a semantically related one. Recent large language models from the GPT family have also been explored. \citet{liu2023deid} proposed DeID-GPT for masking PHI categories and showed that, with zero-shot in-context learning incorporating explicitly HIPAA requirements in the prompts, GPT-4 outperformed fine-tuned transformer models on the same annotated medical texts. 

Text sanitization outside the medical domain includes approaches such as \citet{JUEZHERNANDEZ2023110540}, who propose AGORA, a document de-identification system combined with geoparsing (automatic location extraction from text) using LSTMs and CRFs and trained on Spanish law enforcement data. The authors focus on offering a complete pipeline and location information, while demographic attributes are not part of the information to de-identify. \citet{yermilov-etal-2023-privacy} compared three systems for detecting and pseudonymizing PII: (1) a NER-based one relying on Wikidata; (2) a single-step sequence-to-sequence model trained on a parallel corpus; and (3) a large language model where named entities are first detected using a 1-shot prompt to GPT-3 and then pseudonymized with 1-shot prompts using ChatGPT (GPT-3.5). The authors find that the NER-based approach is best for preserving privacy while LLMs best preserve utility for a text classification and summarization tasks. Finally, \citet{papadopoulou-etal-2022-neural} present an approach to text sanitization, from detection of personal information to privacy risk estimation through the use of language model probabilities, web queries, and a classifier trained on manually labeled data. The present paper builds upon this work. 

\subsection{Privacy-Preserving Data Publishing}
\label{ssec:ppdp}

PPDP approaches to text sanitization rely on a privacy model specifying formal conditions that must be fullfilled to ensure the data can be shared without harm to the privacy of the registered individuals. The most prominent privacy model is $k$-anonymity \citep{Samarati98}, which requires that an individual/entity be indistinguishable from $k$ -1 other individuals/entities. This model was subsequently adapted to text data by approaches such as $k$- safety \citep{chakaravarthy2008efficient} and $k$-confusability \citep{Cumby2011AML}. 

$t$-plausibility \citep{anandan} follows a similar approach, using already detected personal information and ensuring that those are sufficiently generalized to ensure that at least $t$ documents can be mapped to the edited text. \citet{Sanchez2016} present $C$-sanitized, which relies on an information-theoretic approach that computes the point-wise mutual information (using co-occurrence counts from web data) between the person or entity to protect and the terms of the document. Terms whose mutual information ends up above a given threshold are then masked. 

$k$-anonymity was also employed in \citet{bootstrapping} in combination with NLP-based approaches, where based on an assumption of an attacker's knowledge, the optimal set of masking decisions was found to ensure $k$-anonymity. 

Finally, \citet{10.1007/978-3-031-13945-1_12} provided an approach to the evaluation of disclosure risks that relies on training a text classifier to assess the difficulty of inferring the identity of the individual in question based on the sanitized text.

\subsection{Differential Privacy}
\label{ssec:dp}

Differential privacy (DP) is a framework for ensuring the privacy of individuals in datasets \citep{dwork}. It essentially operates by producing randomized responses to queries. The level of artificial noise introduced in each response is optimized such as to provide a guarantee that the amount of information that can be learned about any individual remains under a given threshold. 

\citet{DBLP:conf/post/FernandesDM19} applied DP to text data, in combination with ML techniques by adding noise to the word embeddings of the model. Their work focused on removing stylistic cues from the text as a way to ensure that the author of the text could not be identified by it. \citet{feyisetan2019leveraging} also apply noise to word embeddings in a setting where the geolocation data of an individual is to be protected. 

More recently, \citet{9415242} tried to address the issue of the noise needed for DP causing utility loss in the resulting text by creating duplicates first, and then adding noise, thus reducing the amount of noise needed. \citet{krishna2021adept} sought to address the same issue using an algorithm based on auto-encoders to transform text without losing data utility. Finally, \citet{igamberdiev2023dpbart} introduced DP-BART, a DP rewriting system based on pre-trained BART model, and which seeks to reduce the amount of artificial noise needed to reach a given privacy guarantee.

DP-oriented approaches generally lead to complete transformations of the text, at least for reasonable values of the privacy threshold.  Those approaches are therefore well suited to the generation of synthetic texts, in particular to collect training data for machine learning models. However, they are difficult to apply to text sanitization, as most text sanitization problems are expected to retain the core content of the text and only edit out the personal identifiers. This is particularly the case for court judgments and medical records, as the sanitized documents should not alter the wording and semantic content conveyed in the text.

\section{Datasets}
\label{sec:data}
There only exists a few generic (non-medical) datasets that are devoted to the evaluation of text sanitization approaches. We present below the two datasets used for training, evaluation and error analysis throughout this paper.



\subsection{Text Anonymization Benchmark (TAB)}
\label{ssec:tab}

The TAB corpus \citep{benchmark} is a collection of 1268 European Court of Human Rights (ECHR) court cases, manually annotated to protect the identity of the individual mentioned in the text. These court cases are documents rich in PII that are also freely available to use. The documents are in English and each court case is annotated, among others, with an individual to be protected as well as with detected PII spans containing the following information:

\begin{itemize}
    \item The span's semantic type, according to the 8 categories established in \citet{benchmark}, described in Table \ref{tbl:categories}.
    \item Whether that span needs to be masked in order to protect the privacy of the specified individual, with three possible values: \ENT{DIRECT}, \ENT{QUASI}, or \ENT{NO\_MASK}.
    \item Whether the span is a confidential attribute of the individual, such as religious or philosophical beliefs, political opinions, sexual orientation or sex life, racial or ethnic origin, and health, genetic and biometric data.
    \item Co-reference links between relevant spans that refer to the same entity.
\end{itemize}
\raggedbottom

\begin{table}[t]
\centering
\caption{\label{tbl:categories} Semantic type along with explanations and selected examples}
\begin{tabular}{lp{58mm}p{42mm}}
\textbf{Category} & \textbf{Explanation} & \textbf{Examples}\\
\toprule
\textbf{CODE} & Identification codes and numbers: flight numbers, phone numbers, license plates, etc. & 3086/23, LH3042, 11/14E.2\\
\textbf{ORG} & Names of organizations, such as companies, schools, hospitals, administrations, etc. & Budapest Police Department, Ministry of Justice\\
\textbf{DATETIME} & Description of a specific date, period, time, or duration. & 23 November 2006, 7, 12 and 5 months\\
\textbf{LOC} & Places and locations, such as cities, geographical areas, countries or  addresses. & Austria, Martin County, Belfast\\
\textbf{QUANTITY} & Description of a  quantity, including percentages or monetary values. & 6,932 Ukrainian hryvnyas, two\\
\textbf{PERSON} & Names of people, including nicknames/aliases, usernames, and initials.& Joe Smith, The Rock, YiDA\\
\textbf{DEM} & Demographic attributes of a person, such as nationality, occupation, or education & artist, Italian, MSc in Astrophysics\\
\textbf{MISC} & Every other type of personal information associated to an individual and that does not belong to the categories above. & former patient, high speed car chase, imprisonment\\
\end{tabular}
\end{table}


Example \ref{tab_detect_mask} illustrates a short excerpt where spans underlined in \textit{black} were marked by the annotators, and those underlined in \textit{dark green} were categorized by the annotator as denoting a direct or quasi-identifier to mask. 

\bigskip
\begin{example}
    The case originated in an application (no. \rAB{black}{darkgreen}{44521/04}) against the \rAB{black}{white}{Republic of Poland} lodged with the Court under Article 34 of the Convention for the Protection of Human Rights and Fundamental Freedoms (“the Convention”) by a \rAB{black}{white}{Polish} national, \rAB{black}{darkgreen}{Mr Leszek Kołodziński} (“the applicant”), on \rAB{black}{white}{19 August 2004}.
\label{tab_detect_mask}
\end{example}
\bigskip


The documents are, on average, 1,442 tokens long. The majority of the tokens are labeled as \ENT{QUASI} identifiers (63\%), with fewer tokens being masked as \ENT{DIRECT} identifiers (4.4\%) and the rest of the annotated text spans being left unmasked by the annotators. The majority of the identifiers belonged to the \ENT{DATETIME}, \ENT{ORG}, and \ENT{PERSON} categories which is in line with the domain of the texts.

Some of the court cases (274) were also multi-annotated to allow for evaluation against different solutions since the task is subjective in nature. The inter-annotator agreement on the identifier type (\ENT{DIRECT}, \ENT{QUASI}, \ENT{NO-MASK}), calculated both on the span ($k$ = 0.46) and character level ($k$ = 0.79), shows moderate agreement which is to be expected from a task where multiple correct solutions can exist. We follow the training, development, and test split the authors release with 1,014, 127 and 127 documents respectively.

\subsection{Wikipedia Biographies}
\label{ssec:wiki}

\citet{bootstrapping} released a collection of 553 Wikipedia biographies manually annotated for text anonymization. The annotation task was similar to that of the TAB corpus, except for the absence of annotation for confidential attributes, which was not relevant for this dataset. 

Example \ref{wiki_detect_span} shows a manually annotated text, both for detected PII spans (underlined in \textit{black}), and also for masking decision (underlined in \textit{dark green}). 

\bigskip
\begin{example}
    \rAB{black}{darkgreen}{David Sherwood} is a \rAB{black}{white}{British} \rAB{black}{darkgreen}{tennis coach} and retired \rAB{black}{white}{tennis player}. In his only live \rAB{black}{darkgreen}{Davis Cup match}, \rAB{black}{darkgreen}{Sherwood} played doubles with \rAB{black}{darkgreen}{Andy Murray} beating the \rAB{black}{darkgreen}{Israeli} World No 4 doubles team of \rAB{black}{darkgreen}{Jonathan Erlich} and \rAB{black}{darkgreen}{Andy Ram}.
    \label{wiki_detect_span}
\end{example}
\bigskip

The documents in this dataset are much shorter than those in the TAB corpus. A large majority of the PII spans were marked by the annotators as \ENT{QUASI} identifiers to be masked (56\%) or \ENT{DIRECT} identifiers (14\%), while 30\% of the spans were left as is in the text without needing to be masked, as they were deemed less specific by the annotators and thus, less risky. Most of the identifiers in this dataset belong to the \ENT{DATETIME}, \ENT{ORG}, and \ENT{PERSON} semantic types. 

Of the 553 documents, 22 were also doubly-annotated to account for multiple correct solutions for this task. Cohen’s $k$ calculated both in the span ($k$ = 0.44) and the character level ($k$ = 0.81) on the masking decision showed moderate agreement, highlighting the subjective nature of the task. We follow the split proposed in \citet{olstad-etal-2023-generation}, which includes all the double-annotated documents in the test set (100 documents), while 453 documents are available for training purposes.

\section{Privacy-oriented Entity Recognizer}
\label{sec:er}

Identifying PII spans is the first step in text sanitization. Although many methods rely on some variant of NER, they fail to detect PII spans that are not named entities but are nevertheless (quasi-)identifying.

We detail here our approach to detecting text spans expressing personal information. The approach uses \textit{knowledge graphs} such as Wikidata to create \textit{gazetteers} for specific PII types. Those gazetteers are then combined with a NER model to create a domain-specific silver corpus, which is in turn employed to fine-tune a neural sequence labelling model. This approach to developing a ``privacy-oriented entity recognizer" builds upon earlier work by \citet{papadopoulou-etal-2022-neural}, and provides additional details on various aspects of the gazetteer construction process, model training and empirical evaluation. 


\subsection{Wikidata Properties}
\label{sec:properties}

NER models are, as the term indicates, focused on named entities. However, many instances of the \ENT{DEM} and \ENT{MISC}\footnote{It should be noted that the \ENT{MISC} category employed in this paper does not equate to the \ENT{MISC} category from CoNLL-2003 shared task \citep{tjong-kim-sang-de-meulder-2003-introduction}, which is characterized as a named entity (denoted with a proper name) that is neither a person, organization or place.} categories described in the previous section are not named entities. Examples include someone's occupation, educational background, part of their physical appearance, the manner of their death or an object that is tied to their identity. 

We extract a list of possible values for these two PII categories based on knowledge graphs. In particular, Wikidata\footnote{\url{https://www.wikidata.org}} is a structured knowledge graph containing information in property-value pairs, with a large number of values being adjectives, nouns, or noun phrases. We operated by retrieving all instances of humans from the Wikidata dump file,
 and inspecting Wikidata properties\footnote{\url{https://www.wikidata.org/wiki/Wikidata:Database_reports/List_of_properties/all}} to select those that seems to express either \ENT{DEM} or \ENT{MISC} PII based on their description and their examples.

After filtering, we end up with 44 \ENT{DEM} properties and 196 \ENT{MISC} properties.  Selected examples of each semantic type are shown Table \ref{tbl:example_props}, while a detailed table can be found in Appendix \ref{human_properties}. Some properties were left out due to the high level of false positives they might have introduced if included (e.g. \textit{blood type (P1853)}) or because they mostly contained named entities that would already be detected by a generic NER model.

We then use these properties to traverse the Wikidata instances and save all values into two gazetteers, one for \ENT{DEM} entities\footnote{We also manually add country names and nationalities into the \ENT{DEM} gazetteer to account for cases when the NER failed to detect those and the gazetteer lacked this information.} and one for \ENT{MISC} entities.

\begin{table}[t]
\caption{\label{tbl:example_props} Selected examples of Wikidata properties of type \ENT{DEM} or \ENT{MISC}.}
\begin{tabular}{*4l}
\textbf{Property} & \textbf{ID} & \textbf{Examples} & \textbf{Type}\\
\toprule
hairstyle/hairlength & P8839 & baldness, comb over, curls  & DEM \\
religion or worldview & P140 & agnosticism, occultism, free-thought & DEM \\
sex or gender & P21 & agender, queer, non-binary & DEM \\
position held & P39 & U.S. senator, Mayor of Lopigna, chief physician & DEM \\
\midrule
vessel & P1876 & Soyuz-TMA, galley, ferry ship & MISC \\
convicted of & P1399 & forgery, harassment, conspiracy to murder & MISC \\
manner of death & P1196 & capital punishment, disease, matricide & MISC \\
genre & P136 & metalcore, war film, icelandic hip hop & MISC \\
\end{tabular}
\end{table}

\subsection{Silver Corpus and Model Fine-tuning}
\label{sec:corpus}

A silver corpus of 5000 documents is then compiled, consisting in our experiments with the datasets of Section \ref{sec:data} of 2500 European Court of Human Rights cases and 2500 Wikipedia summaries \citep{lebret2016neural}. To automatically label the documents, we first run a generic NER model\footnote{We used here a RoBERTa model fine-tuned on the Ontonotes v5 corpus using  \href{https://spacy.io/}{spaCy}'s implementation.} to detect named entities. We then apply the two \ENT{DEM} and \ENT{MISC} gazetteers and tag each match with their respective label. In case of overlap, we keep the longest span, e.g. keep ``Bachelor in Computer Science" instead of ``Bachelor" and ``Computer Science" as two separate spans. 

We then employ this silver corpus to fine-tune a RoBERTa \citep{DBLP:journals/corr/abs-1907-11692} model, thus creating a \textit{privacy-oriented entity recognizer}. Detailed training parameters can be found in Table \ref{tbl:Params} in Appendix \ref{parameters}.

\subsection{Evaluation}
\label{ssec:er_results}

The evaluation results of the privacy-oriented entity recognizer are shown in Table \ref{tbl:perf1}. The total precision, recall and $F_1$ scores are provided in two versions: one where we take label types into account and one where we do not. In the latter case, we thus only consider whether the token was marked as PII or not\footnote{This differs from an earlier evaluation in \citet{papadopoulou-etal-2022-neural} where only a subset of labels such as \ENT{ORG} and \ENT{LOC} or \ENT{MISC} and \ENT{QUANTITY} were considered interchangeable.}. The evaluation is conducted on the test sets of the TAB corpus and the Wikipedia collection of biographies. As the documents from those test sets were annotated by several annotators, the results are calculated using a micro-average over all annotators.

\begin{table}[t!]
\centering
\caption{\label{tbl:perf1} Token-level precision (\textit{P}), recall (\textit{R}) and $F_1$ score per semantic type on the test sets of the Wikipedia biographies and TAB corpus. We also report micro-averaged performance scores under two conditions: one where we require exact matches on the predicted label, and one where we only distinguish between PII-tokens and non-PII-tokens (thus conflating all PII types into one group).}
\begin{tabular}{lp{10mm}p{10mm}p{10mm}|p{10mm}p{10mm}p{10mm}}
\toprule
    & \multicolumn{3}{c|}{\textbf{Wikipedia}} & \multicolumn{3}{c}{\textbf{TAB}} \\
   & \textbf{ P } &   \textbf{R}  &   \textbf{$F_1$}  &   \textbf{ P } &   \textbf{R}  &   \textbf{$F_1$}  \\
\midrule
\ENT{\textbf{CODE}}   &  \textit{n/a}  & \textit{n/a} & \textit{ n/a} &  0.98  & 0.95  &    0.97  \\
\ENT{\textbf{PERSON}}   & 0.97   &  0.88  &  0.92  &  0.98  &   0.94 &   0.96   \\
\ENT{\textbf{ORG}}   &  0.86  &  0.79  &   0.82 &  0.75  &  0.81  &    0.78  \\
\ENT{\textbf{LOC}}   &  0.86  &  0.79  &  0.82  &  0.51  &  0.80  &   0.62   \\
\ENT{\textbf{QUANTITY}}   &  0.49  &  0.83  &  0.62  &  0.39  &  0.82  &   0.53   \\
\ENT{\textbf{DATETIME}}   & 0.84   &  0.96  & 0.90  &  0.92 & 0.99   &    0.95  \\
\ENT{\textbf{DEM}}   &  0.56  &  0.84  &  0.67  &  0.10  &  0.47  &    0.16  \\
\ENT{\textbf{MISC}}   &  0.52  &  0.61  &  0.56  &  0.03  &   0.23  &   0.06   \\
\midrule
\midrule
\textbf{Micro-avg. w/ label match} &  0.72  &  0.81  &  0.76  &  0.57  &   0.84 &   0.68  \\
\textbf{Micro-avg. w/o label match} &  0.83  & 0.92   &  0.87  &  0.63  &  0.93  &  0.75    \\
\end{tabular}
\end{table}

We notice a clear difference between the performance with and without label match. This indicates a label disagreement between the model and the gold annotations, in particular for categories such as \ENT{ORG} and \ENT{LOC}. This does not, however, affect the detection of the text spans themselves. We also observe that the model performs better for the Wikipedia biographies than TAB corpus, indicating that the PII in Wikipedia are easier to detect than in TAB. 


Breaking down the results based on semantic type, we observe that \ENT{CODE}, \ENT{PERSON}, \ENT{ORG}, and \ENT{DATETIME} seem to consistently exhibit strong results on the two datasets. This is likely due to those categories being easier to circumscribe and recognize from surface cues. In contrast, \ENT{DEM} and \ENT{MISC}, as well as \ENT{QUANTITY} are harder to detect, as those encompass a larger set of possible spans, include many which are not named entities. 


In the annotated collection of Wikipedia biographies, \ENT{DEM} spans are maximum 9 tokens long, with an average of 1.3 tokens, while for \ENT{MISC} the maximum number of tokens is 42, with an average of 2.3 tokens. For TAB, the maximum number of tokens is much higher. For \ENT{DEM} it is 24, while for \ENT{MISC} it is 785 (a long transcript), with an average of 1.6 and 4 respectively. On the silver corpus, the label outputs from the model for \ENT{DEM} and \ENT{MISC} are at maximum 11 and 26 tokens respectively, while on average they are 1.1 tokens long, being much closer to Wikipedia standards than TAB ones.

\subsubsection*{Examples}

We provide below two examples of text spans from the two corpora, comparing PII spans annotated by a human annotator (\textit{black} line) with the ones detected the generic NER model (\textit{red} line) and the privacy-oriented entity recognizer (\textit{dark green} line).

\bigskip
\begin{example}
    \uuuline{Percy Parke Lewis} \uuuline{(1885-1962)} was an \uuuline{American} \rAB{black}{darkgreen}{architect}.
\label{example_er}
\end{example}
\bigskip
In the above example, the privacy-oriented recognizer detected the demographic attribute \textit{architect}, in line with the human annotation, while it was ignored from the standard NER model.  

\bigskip
\begin{example}
    On \uuuline{1 December 2005} the applicant was arrested and remanded in custody. In \uuuline{2006} he was convicted of \rAB{red}{darkgreen}{three} counts of \rAB{darkgreen}{white}{battery} and \rAB{darkgreen}{white}{robbery} and received \rAB{darkgreen}{white}{prison} sentences ranging \uuuline{from two to four years}.
\label{tab_er}
\end{example}
\bigskip

We observe in the above example that the privacy-oriented entity recognizer also labeled \textit{battery} and \textit{robbery} as reasons of conviction (\ENT{MISC}) as well as \textit{prison} (\ENT{MISC}).


\subsection{Label Disagreement}
\label{ssec:ld}

As mentioned in Section \ref{sec:corpus}, Wikidata properties were used to annotate entity types in the text to train the entity-recognition model. Wikidata pages are either generated when a corresponding Wikipedia article is created or they are set up manually by human editors. Both of these make Wikidata pages prone to possible inaccuracies, something we encountered while performing an error analysis on the output of the model when applied to the datasets.

Two cases can be distinguished when there is disagreement between the expert annotations and the model outputs: either (1) both system and gold labels can be considered correct or (2) the system made an error, either as a false positive or false negative. We discuss below some of those disagreements.

The pair \ENT{LOC} and \ENT{ORG} are often confused with one another, which is an error commonly found in many NER systems as well. In Example \ref{gurajat} below, the underlined word was labeled as a \ENT{LOC} by the annotator but the predicted label was \ENT{ORG}.
\bigskip
\begin{example}
    \textit{"... in the Government of \underline{Gujarat"}.}
\label{gurajat}
\end{example}
\bigskip
The \ENT{QUANTITY}-\ENT{MISC} and \ENT{MISC}-\ENT{DEM} label pairs are also often confused:
\bigskip
\begin{example}
    \textit{"... and sentenced to \underline{life imprisonment}."}
\label{imprisonment}
\end{example}
\bigskip
\begin{example}
    \textit{"Quinn \underline{co-founded} Crash Override."}
\label{override}
\end{example}
\bigskip
\begin{example}
    \textit{"... was suffering from \underline{depression and memory loss} ..."}
\label{depression}
\end{example}
\bigskip

In Example \ref{imprisonment}, the underlined span was labeled as \ENT{QUANTITY} by the annotator, potentially interpreting the span as a time duration. The entity recognizer, on the other hand, labeled the same span as \ENT{MISC} due to properties such as \textit{penalty} or \textit{cause of death} that were used to form the \ENT{MISC} gazetteer. 
Similarly, in Examples \ref{override} and \ref{depression}, the human annotators decided to mark these spans as \ENT{DEM}, viewing job titles and diagnosis are demographic attributes. The entity recognizer detected both spans and labeled them as \ENT{MISC}, since, medical related properties or \textit{cause of death} and properties containing job related words that are not nouns were used for the \ENT{MISC} gazetteer.

Figure \ref{fig:label_graph} in Appendix \ref{label_confusion} shows a detailed breakdown of all pairs of label disagreements, common between the two datasets. The \ENT{MISC} semantic type is often found in many cases of label disagreements. In the following section we make an effort to analyze this semantic type in more detail.

\subsection{MISC Semantic Type}



The \ENT{MISC} category was defined in Section \ref{sec:data} as a type of personal information that cannot be assigned to any of the other categories, namely \PERSON{}, \ORG{}, \LOC{}, \CODE{}, \DATETIME{}, \QUANTITY{} or \DEM{}. We conducted a qualitative analysis of this category in both the TAB corpus and the collection of Wikipedia biographies to better understand the coverage of this PII type. 

We found that most \MISC{} spans could be mapped to the following 6 sub-categories:
\begin{itemize}
    \item[] \ENT{EVENTS}: four-man, mixed doubles, Second World War
    \item[] \ENT{QUOTES}: ``the need for further exploration of your insight into, and responsibility for, the index offence and the apparent lack of empathy towards the victim"
    \item[] \ENT{HEALTH}: bullet entry hole on the face, multiple sclerosis, blood poisoning
    \item[] \ENT{WORKS OF ART}: (\textit{only in Wikipedia}) Star Trek: The Original Series, The Book and the Brotherhood 
    \item[] \ENT{OFFENSES}: (\textit{only in TAB}) tax asset stripping, seriously injured, a charge of attempted murder
    \item[] \ENT{LAWS}: (\textit{only in TAB}) sections 1 and 15 of the Theft Act 1968, Article 125 of the Criminal Code, 19 § 4
\end{itemize}

Some of those categories are quite general and can be found in a large number of text domains, such as \ENT{EVENTS} or \ENT{HEALTH}. Others are specific to a text domain, such as \ENT{OFFENSES}, which are most likely to be observed in court cases. 

Finally, there were also examples of \ENT{MISC} that could not be grouped into the 6 categories above. These include, for example, various types of objects or occupation related words, or anything else that is difficult to assign to a specific group:

\begin{itemize}
    \item[] \ENT{ITEM}: gold metal, weekly newspaper, car, motorcycle, police vehicle
    \item[] \ENT{OCCUPATION RELATED}: science policy, basketball skills, captained, coached
    \item[] \ENT{OTHER}: empowering the backward classes, responsibility for child poverty and childcare, premeditated murder [assassinat] and in the alternative with murder [meurtre]
\end{itemize}

The above analysis is constrained to the two datasets employed in this paper. A privacy-oriented entity recognizer may, of course, be enriched with other sub-categories. As pointed out by \citet{benchmark}, \ENT{MISC} was also the PII category human annotators found most difficult to agree on. Despite this difficulty, recognizing \ENT{MISC} spans remains an important part of text sanitization, as many of those text spans do provide detailed information about the individual in question.



\section{Privacy Risk Indicators}
\label{sec:indicators}

Many text sanitization approaches simply operate by masking all detected PII spans. This may, however, lead to overmasking, as the actual risk of re-identification may vary greatly from one span to another. In many documents, a substantial fraction of the detected text spans may be kept in clear text without notably increasing the re-identification risk. For instance, in the TAB corpus, only 4.4 \% of the entities were marked by the annotators as direct identifiers, and 64.4 \% as quasi-identifiers, thus leaving 31.2 \% of the entities in clear text. To assess which text spans should be masked, we need to design privacy risk indicators able to determine which text spans (or combination of text spans) actually constitute a re-identification risk. 

We present 5 possible approaches for inferring the re-identification risk associated with text spans in a document. Those 5 indicators are respectively based on:
\begin{enumerate}
\item LLM probabilities, 
\item Span classification,
\item Perturbations,
\item Sequence labelling,
\item Web search.
\end{enumerate}

The web search approach can be applied in a zero-short manner without any fine-tuning. The two methods respectively based on LLM probabilities and perturbations require a small number of labeled examples to adjust a classification threshold or fit a simple binary classification model. Finally, the span classification and sequence labelling approaches operate by fine-tuning an existing language model, and are thus the methods that are most dependent on getting a sufficient amount of labeled training data to reach peak performance. This training data will typically take the form of human decisions to mask or keep in clear text a given text span. 

We present each method in turn and provide an evaluation and discussion of their relative benefits and limitations in Section \ref{sec:eval}.


\subsection{LLM Probabilities}
\label{ssec:tabular}

The probability of a span as predicted by a language model is inversely correlated with its informativeness: a text span that is harder to predict is more informative/surprising than one that the language model can easily infer from the context \citep{zarcone2016salience}. 
The underlying intuition is that text spans that are highly informative/surprising are also associated with a high re-identification risk, as they often correspond to specific names, dates or codes that cannot be predicted from the context. 


Concretely, we calculate the probability of each detected PII span given the entire context of the text by masking all the (sub)words of the span, and returning a list with all log-probabilities (one per token) referring to the span as calculated by a large, bidirectional language model, in our case BERT (large, cased) \citep{devlin-etal-2019-bert}. Those probabilities are then aggregated and employed as features for a binary classifier that outputs the probability of the text span being masked by a human annotator.


The list of log-probabilities for each span has a different length depending on the number of tokens comprised in the text span. We therefore aggregate this list by reducing it to 5 features, namely the \textit{minimum}, \textit{maximum}, \textit{median} and \textit{mean} log-probability as well as the \textit{sum} of log-probabilities in the list. In addition, we also include as feature an encoding of the PII type assigned to the span by the human annotators (see Table \ref{tbl:categories}).

For the classifier itself, we conduct experiments using a simple logistic regression model as well as a more advanced classification framework based on AutoML. AutoML \citep{he2021automl} provides a principled approach to hyper-parameter tuning and model selection, efficiently searching for the hyper-parameters and model combination that yield the best performance. We use the Autogluon toolkit \citep{autogluon} in our experiments, and more specifically AutoGluon's Tabular predictor \citep{autogluon}, which consists of an ordered sequence of base models. After each model is trained and hyper-parameter optimization is performed on each, a weighted ensemble model is trained using a stacking technique.  
A list of all base models of the Tabular predictor is shown in Table \ref{tbl:tabular_models}, in Appendix \ref{model}.
We use the training splits of the Wikipedia biographies and the TAB corpus to fit the classifier. 



\subsection{Span Classification}
\label{ssec:multimodal}

The approach in the previous section has the benefit of being relatively general, as it only considers the aggregated LLM probabilities and predicted PII type. However, it does not take into account the text content of the span itself. We now present an indicator that includes this text information to predict whether a text span constitutes a high privacy risk, and should therefore be masked.

Concretely, we enrich the classifier from the previous section by adding to the features a vector representation of the text span derived from a large language model. The classifier thus relies on a combination of numerical features (the aggregated LLM probabilities discussed above, to which we add the number of words and subwords of the span), a categorical feature (an encoding of the PII type), and a vector representation of the text span. Crucially, the transformer-based language model that produces this vector representation is fine-tuned to this prediction task. 

\begin{figure}[t]
\centering
\includegraphics[width = 0.95\textwidth]{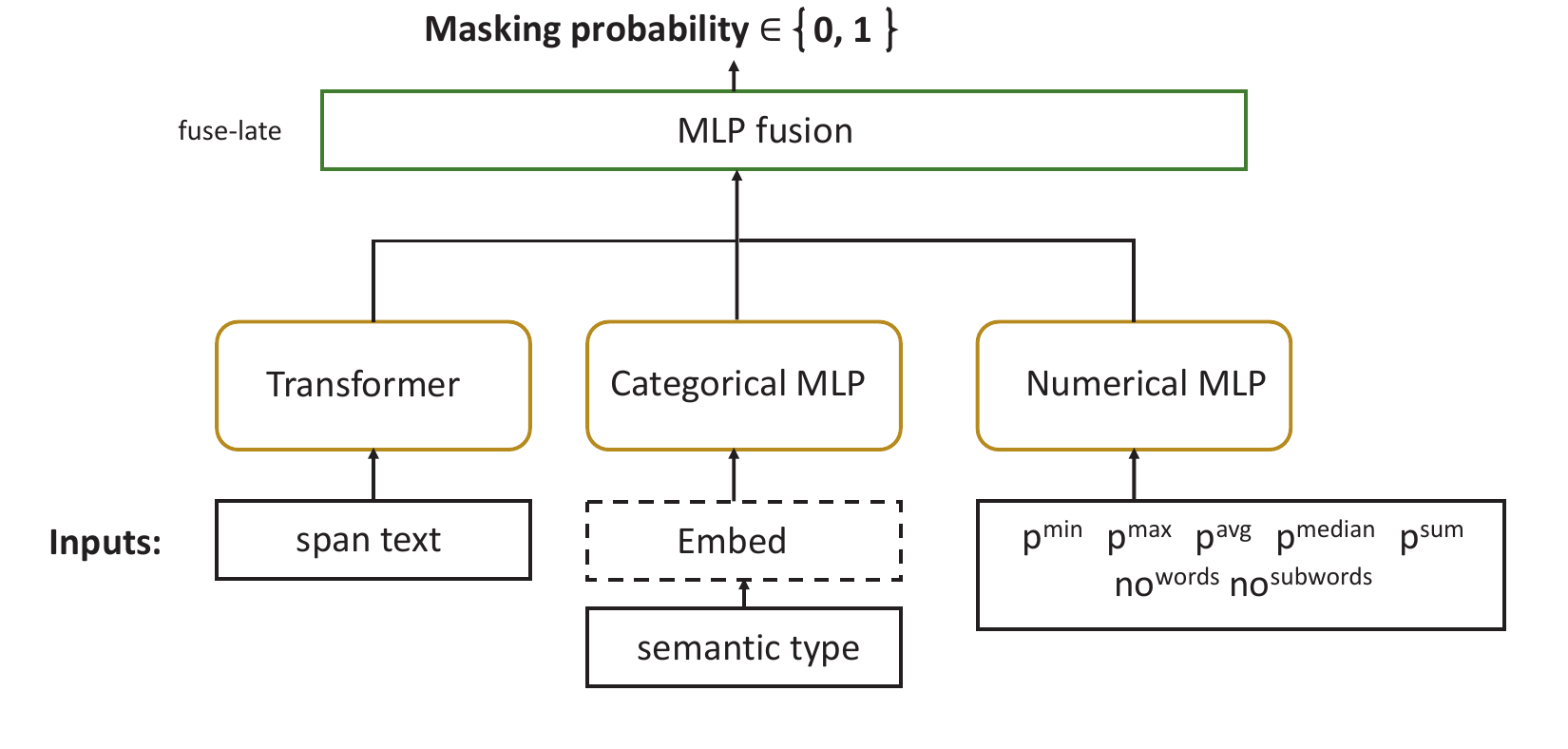}
\caption{Overview of the model architecture for span classification, including both numerical, text and categorical features. Adapted from \citet{multimodal}.}
\label{fig:multimodal_predictor}
\end{figure}

The resulting model architecture is illustrated in Figure \ref{fig:multimodal_predictor}. Concretely, the model training is done using AutoGluon's Multimodal predictor \citep{multimodal}, making it easier to incorporate a wide range of feature types, including textual content processed with a neural language model. We employ here the Electra Discriminator model due to its good encoding performance \citep{DBLP:conf/iclr/ClarkLLM20,multimodal}.  The categorical and numerical features are processed by a standard MLP. After each model is trained separately, their output is pooled (concatenation) by a MLP model using a fuse-late strategy near the output layer. Table \ref{tab:multimodal_features} provides some examples of input. 

\begin{table*}[ht]
      \caption{Selected examples of input features for the span classification model. The aggregated LLM probabilities are in logarithmic form. \textbf{nb}\textsuperscript{w} and \textbf{nb}\textsuperscript{sw} respectively correspond to the number of words and subwords (token) in the span.}
      \begin{footnotesize}
     \begin{tabular}{lllllllll|l}
         \toprule
         \textbf{span text}& \textbf{PII type} &  
 \textbf{p}\textsuperscript{min} &\textbf{p}\textsuperscript{max} & \textbf{p}\textsuperscript{avg} & \textbf{p}\textsuperscript{mdn} & \textbf{p}\textsuperscript{sum} & \textbf{nb}\textsuperscript{w} & \textbf{nb}\textsuperscript{sw} &\textbf{label} \\
         \midrule
          36244/06 & \CODE{} &  -5.7 & -1.9 & -3.7 & -3.3 & -18.4 & 3 & 2 & \ENT{MASK} \\
         Kingdom of Denmark & \ORG{} &  -4.3 & -3.9 & -4.1 & -4.2 & -12.4 & 3 & 0 & \ENT{NO\_MASK} \\
          winger & \MISC{} & -1.5 & -1.5 & -1.5 & -1.5 & -1.5 & 1 & 0 & \ENT{NO\_MASK} \\
          Reniers &  \PERSON{} &-8.6 & -4.2 & -6.4 & -6.4 & -12.8 & 1 & 1 & \ENT{MASK} \\
         
     \end{tabular}
     \end{footnotesize}
     \label{tab:multimodal_features}
 \end{table*}

For our experiments, we again fit the classifier using the training datasets of the annotated collection of Wikipedia biographies and the TAB corpus. 

\subsection{Perturbations}
\label{ssec:petrubation}

The two previous risk indicators use features extracted from the text span to emulate masking decisions from human annotators. They are not, however, able to quantify how much information a PII span can contribute to the re-identification. They also consider each PII span in isolation, thereby ignoring the fact that quasi-identifiers constitute a privacy risk precisely because of their combination with one another. We now turn to a privacy risk indicator that seeks to address those two challenges.


This approach is inspired by saliency methods \citep{li-etal-2016-visualizing,ding-etal-2019-saliency} used to explain and interpret the outputs of large language models. One such method is \textit{perturbation} when one changes the input to the model by either modification or removal, and observes how this change affects the downstream predictions \citep{Kindermans2019,li2017understanding, serrano-smith-2019-attention}. In our case, we \textit{perturb} the input by altering one or more PII spans and analyse the consequences of this change on the probability produced by the language model for another PII span in the text.

Consider the following text:
\bigskip
\begin{example}
    ``Michael Linder is an \underline{American} \dashuline{television producer} and broadcast journalist based in Los Angeles [...] In 1995-6 Linder created and produced ``Berserkistan" [..] From 2004 through 2009, Linder served as a radio journalist for KNX-AM. [...]"
\label{linder}
\end{example}
\bigskip
The target span \underline{American} has a log probability of -0.006 given the rest of the document. However, if the span ``\dashuline{television producer}'' ends up being masked, its log probability becomes -2.062, indicating that the latter is informative to the target. 

To assess which spans to mask, we use a RoBERTa large model \citep{DBLP:journals/corr/abs-1907-11692} and all possible combinations of PII in the text in a perturbation-based setting. We begin by calculating the probability of a target span with the rest of the context available in the text. We then mask in turn every other PII span in the document and recompute the probability of the target span to determine how the absence of the PII span affects the probability of the target span.

To obtain sensible estimates, we must mask all spans that refer to the same entity while calculating the probabilities. In Example \ref{linder}, \textit{Michael Linder} and \textit{Linder} refer to the same entity. We use in our experiments the co-reference links already annotated in the TAB corpus and the collection of Wikipedia biographies.



Based on the difference between the original and the new target probability, we select the PII spans that make the target span more difficult to predict. To determine a threshold for deciding which spans constitute an unacceptable risk, we devise an objective function based on the privacy-oriented precision and recall metrics outlined in Section \ref{sec:metrics}. More precisely, assuming that PII spans are masked when the difference between the original and new target probability is greater than a threshold $t$, we look for the optimal threshold value $t^*$ that maximises the sum of three scores, namely the precision $p^t$, recall on direct identifiers $r_{\textrm{direct}}^t$ and recall on quasi identifiers $r_{\textrm{quasi}}^t$:
\begin{equation}
t^* = \argmax_t \left(p^t + r_{\textrm{direct}}^t + r_{\textrm{quasi}}^t\right)
\label{cost_function}
\end{equation}

The three above scores $p^t$, $r_{\textrm{direct}}^t$ and $r_{\textrm{quasi}}^t$ are obtained using a threshold $t$ on the difference of probabilities to decide which spans to mask. Those scores are computed in our experiments using the training set of our corpora. The resulting thresholds are different for the Wikipedia biographies (approximately 3.5) and the TAB corpus (approximately 10). Figure \ref{fig:perturbation_graph} in Appendix \ref{per_graphs} shows the performance of this risk indicator at different thresholds for both datasets. 




\subsection{Sequence Labelling}
\label{ssec:supervised}

Yet another approach to indirectly assess the re-identification risk based on masking decisions from experts is to estimate a sequence labelling model. Compared to the previous methods, this method is the one that is most dependent on the availability of in-domain, labeled training data.


For this approach, we fine-tune a encoder-type language model on a token classification objective, each token being assigned to either \ENT{MASK} or \ENT{NO\_MASK}. For the Wikipedia biographies, we rely on a RoBERTa model \citep{DBLP:journals/corr/abs-1907-11692}, while we switch to a Longformer model \citep{beltagy2020longformer} for TAB given the length of the court cases, as proposed in \citet{benchmark}. Due to discrepancies between the manually labeled spans or detected by the privacy-oriented entity recognizer, and the ones created by the fine-tuned model, we operate under two possible setups:
\begin{itemize}
    \item Full match: We assume that a span constitutes a high re-identification risk if all of its tokens are marked as \ENT{MASK} by the fine-tuned Longformer/RoBERTa.  
    \item Partial match: We consider that the span has a high risk if at least one token is marked as \ENT{MASK} by the Longformer/RoBERTa model.  
\end{itemize}

\subsection{Web Search}
\label{ssec:wqra}

Web query results can also be used to approximate the re-identification risk. This approximation relies on the assumption that text spans with an inadmissible level of re-identification risk will have web results that mention the individual we are trying to protect. 
Building from earlier work in \citet{papadopoulou-etal-2022-neural}, we first query the person's name to obtain URLs related to that person. We then query the PII spans of the document. For each span, we then compare the resulting URLs with those from querying the search engine with the person's name. If the intersection of these two sets of URLs is not empty, the text span is considered to pose a high re-identification risk. 
More formally, whether a text span poses a risk of re-identification, $R_{web}$, is defined as:

\begin{equation}
    R_{web}=\begin{cases}
    0, &\text{ if } \{URLs_{e}\}\cap\{URLs_{i}\} = \emptyset \\
    1, &\text{ otherwise } 
    \end{cases}
    \label{eq:web_risk}
\end{equation}

where $\{URLs_{e}\}$ is the set of URLs resulting from querying the web search engine with the text span and $\{URLs_{i}\}$ is the set of URLs mentioning the person to protect. 

We used the Google API to query for each target individual in a given document and the unique text spans that occur in a given document\footnote{Web searches are from the period spanning July 2023 to September 2023.}.
The Google API provides 10 results per page. We limit the experiment to the top 20 results (i.e. first two pages from the web search). To avoid a prohibitively high number of API calls, we also constrain the search to individual text spans, although the same approach can in principle be extended to combinations of PII spans.


We also used the total number of hits reported by the Google search API for each PII span query. The assumption here is that if a search yields a larger number of responses, there is a higher chance that one of those responses will contain information about the target individual. However, generic search queries are also likely to return many responses. 
Therefore we considered applying an upper and lower bound on the total number hits. These thresholds were set experimentally to maximize the token-level $F_1$ scores on the TAB development set. This resulted in a lower limit of 100 hits and no upper limit. This method is limited by the potential unreliable nature of the total responses reported by web search engines, as shown in \citet{sanchez2018survey}.


\section{Analysis of Privacy Risk Indicators}
\label{sec:eval}

We now evaluate the privacy risk indicators detailed in the previous section, both in isolation (Section \ref{subsec:eval}) and in combination (Section \ref{ssec:combos}) and discuss the main findings.

\subsection{Evaluation Metrics}
\label{sec:metrics}

The evaluation of text sanitization must strike a balance between two competing considerations:
\begin{itemize}
    \item \textit{Privacy risk}: text spans that have a high risk of leading, directly or indirectly, to the re-identification of the individual should be masked.
    \item \textit{Data utility}: the sanitized text should preserve as much semantic content as possible.
\end{itemize}

The traditional approach to evaluating text sanitization is to rely on a manually labeled set of documents and use metrics such as precision, recall and $F_1$ score to measure the overlap between the human decisions and the system outputs. The recall is associated with the privacy risk, as a high recall indicates that most of the PII spans that should have been masked, are indeed masked. Similarly, the precision is correlated with the data utility, as a high precision expresses that the resulting sanitization did not mask unnecessary text spans.  


As argued in \citet{benchmark}, the use of standard precision and recall measures has, however, a number of important shortcomings. They do not capture the fact that some identifiers have a more important influence on the privacy risk than others (for instance, omitting to mask a full person name is a more serious threat than omitting a vague quasi-identifier). They also ignore the fact that a personal identifier is only protected if all its occurrences are masked: if a person name is mentioned 4 times in a document, and 3 of those occurrences are masked, the re-identification risk remains unchanged, as the identifier remains available in clear text in one occurrence. Finally, text sanitization may admit several, equally correct solutions. 

To provide a more fine-grained assessment of the text sanitization quality, \citet{benchmark} present a set of three privacy-oriented metrics, micro-averaged over multiple annotators: 
\begin{description}
\item \textbf{Entity recall on direct identifiers} Micro-averaged, entity-level recall score calculated only for direct identifiers. If an entity is mentioned multiple times in the text, all mentions must to be masked for it to be considered correct. 
\item \textbf{Entity recall on quasi identifiers} Micro-averaged, entity level recall score calculated only for quasi identifiers. In case of multiple mentions of the same entity, all mentions must again be masked. 
\item \textbf{Weighted Precision} Micro-averaged, token-level precision where each token is weighted by its information content (measured using a language model such as BERT). 
\end{description}

The evaluation results below are computed according to those three metrics.

\subsection{Experimental Results}
\label{subsec:eval}

 We first evaluate the five privacy risk indicators using PII spans manually labeled by human annotators from both the TAB corpus and the Wikipedia biographies. The results are shown in Table \ref{tbl:manual}. The approaches based on LLM probabilities and span classification were trained on the corresponding training set (either Wikipedia or TAB), and the threshold of the perturbation-based method was similarly adjusted from the training set. The web search results rely on the two alternatives techniques presented in Section \ref{ssec:wqra}, respectively based on the intersection of URLs or the estimated number of hits. We also provide the results obtained with a simple baseline that masks all PII spans.

We then evaluate the risk indicators with the PII spans actually detected by the privacy-oriented entity recognizer described in Section \ref{sec:er}, thereby enabling us to assess the end-to-end performance of the proposed approaches. The results are shown in Table \ref{tbl:automatic}. Note that, in this setup, the majority rule baseline (which masks all PII spans) leads to recall scores lower than 1 due to detection errors arising from the privacy-oriented entity recognizer. 


\begin{table*}[t]
\begin{center}
\caption{\label{tbl:manual}  Evaluation of the privacy risk indicators with manually labeled text spans. The results are obtained by comparing the text spans predicted as risky by each method against the masking decisions of human experts. We report \textit{token-level} and \textit{entity-level} scores. \textbf{P} denotes the token-level precision, $\mathbf{P_{w}}$ the token-level precision weighted by information content (see \citet{benchmark} for details), $\mathbf{R}_{\textrm{all}}$ the token-level recall of all personal identifiers, and $\mathbf{F_1}$ the token-level $F_1$ score on all identifiers. $\mathbf{R}_{\textrm{ent}}$ denotes the entity-level recall for all identifiers, while $\mathbf{R}_{\textrm{direct}}$ the entity-level recall of direct identifiers and $\mathbf{R}_{\textrm{quasi}}$ the entity-level recall of quasi-identifiers. The results for a given document are micro-averaged over all annotators.}
\vspace{3mm}
\begin{footnotesize}
\begin{tabular}{lrrrr:rrr}
\toprule
  &\hspace{-2mm}\textbf{P} & $\mathbf{P_{w}}$ & $\mathbf{R}_{\textrm{all}}$ &\hspace{-2mm}$\mathbf{F_1}$ &\hspace{-3mm}$\mathbf{R}_{\textrm{ent}}$ &\hspace{-3mm}$\mathbf{R}_{\textrm{direct}}$ &\hspace{-3mm}$\mathbf{R}_{\textrm{quasi}}$  \\
 \midrule
 \multicolumn{8}{c}{\textbf{Wikipedia biographies (test set)}}\\
  \midrule
Majority rule (mask everything) & 0.80 & 0.84 & \textbf{1.00} & 0.89 & \textbf{1.00} & \textbf{1.00} & \textbf{1.00}  \\[1mm]\hdashline\\[-2mm]
LLM probabilities (logistic regression)  & 0.87 & 0.89 & 0.92 & 0.89 & 0.85 & \textbf{1.00} & 0.84 \\
LLM probabilities (AutoML model) & 0.89 & 0.91 & 0.93 & \textbf{0.91} & 0.89 & \textbf{1.00} & 0.88 \\[1mm]\hdashline\\[-2mm]
Span classification (AutoML model) & 0.93 & 0.94 & 0.87 & 0.90 & 0.82 & 0.99 & 0.81 \\[1mm]\hdashline\\[-2mm]
Perturbations (threshold $\approx 3.5$) & 0.80 & 0.84 & \textbf{1.00} & 0.89 & \textbf{1.00} & 0.99 & \textbf{1.00} \\[1mm]\hdashline\\[-2mm]
Sequence labelling (RoBERTa, strict match) & 0.84 & 0.89 & 0.81 & 0.82 & 0.78 & 0.85 & 0.77\\
Sequence labelling (RoBERTa, partial match) & 0.83 & 0.86 & 0.97 & 0.89 & 0.97 & \textbf{1.00} & 0.96\\[1mm]\hdashline\\[-2mm]
Web search (URL $\cap$) & \textbf{0.98} & \textbf{0.99}  & 0.22 & 0.36 & 0.08 & 0.52 & 0.03  \\
Web search (\# of hits) & 0.80 & 0.84 & 0.99 & 0.88 & 0.99 & 0.95 & 0.99 \\[1mm]
 \midrule
 \multicolumn{8}{c}{\textbf{Text Anonymization Benchmark (test set)}}\\
   \midrule
Majority rule (mask everything) & 0.52 & 0.56 & 1.00 & 0.69 & \textbf{1.00} & \textbf{1.00} & \textbf{1.00}  \\[1mm]\hdashline\\[-2mm]
LLM probabilities (logistic regression)  & 0.53 & 0.57 & 0.96 & 0.68 & 0.93 & 0.91 & 0.93\\
LLM probabilities (AutoML model) & 0.61 & 0.64 & 0.93 & 0.73 & 0.93 & \textbf{1.00} & 0.92\\[1mm]\hdashline\\[-2mm]
Span classification (AutoML model) & 0.74 & 0.77 & 0.92 & 0.83 & 0.93 & \textbf{1.00} & 0.92 \\[1mm]\hdashline\\[-2mm]
Perturbations (threshold $\approx 10)$ & 0.52 & 0.56 & \textbf{1.00} & 0.69 & \textbf{1.00} & \textbf{1.00} & \textbf{1.00}\\[1mm]\hdashline\\[-2mm]
Sequence labelling (Longformer, strict match) & 0.90 & 0.91 & 0.87 & 0.88 & 0.88 & 0.96 & 0.87\\
Sequence labelling (Longformer, partial match) & 0.87 & 0.88 & 0.95 & \textbf{0.90} & 0.95 & \textbf{1.00} & 0.95 \\[1mm]\hdashline\\[-2mm]
Web search (URL $\cap$) & \textbf{0.95} &  \textbf{0.96} & 0.13 & 0.22 & 0.04 & 0.41 & 0.01 \\
Web search (\# of hits) & 0.52 & 0.56 & 0.98 & 0.68 & 0.97 & 0.93 & 0.98\\
\end{tabular}
\end{footnotesize}
\end{center}
\bigskip
\end{table*}

\begin{table*}[t]
\begin{center}
\caption{\label{tbl:automatic}  Evaluation of the privacy risk indicators with PII spans detected by the privacy-oriented entity recognizer of Section \ref{sec:er}, using the same metrics as in Table \ref{tbl:manual}.}
\begin{footnotesize}
\begin{tabular}{lrrrrrrr}
\toprule
  & \textbf{P} &\hspace{-2mm}$\mathbf{P_{w}}$ & $\mathbf{R}_{\textrm{all}}$ & \hspace{-2mm}$\mathbf{F_1}$&\hspace{-3mm}$\mathbf{R}_{\textrm{ent}}$&\hspace{-3mm}$\mathbf{R}_{\textrm{direct}}$&\hspace{-3mm}$\mathbf{R}_{\textrm{quasi}}$\\
 \midrule
 \multicolumn{8}{c}{\textbf{Wikipedia biographies (test set)}}\\
  \midrule
  Majority rule (mask everything) & 0.66 & 0.75 & \textbf{0.93} & 0.77 & \textbf{0.91} & 0.88 & \textbf{0.91}  \\[1mm]\hdashline\\[-2mm]
LLM probabilities (logistic regression)  & 0.82 & 0.86 & 0.83 & 0.82 & 0.75 & 0.87 & 0.74 \\
LLM probabilities (AutoML model) & 0.83 & 0.87 & 0.84 & \textbf{0.84} & 0.78 & 0.87 & 0.78 \\[1mm]\hdashline\\[-2mm]
Span classification (AutoML model) & 0.87 & 0.90 & 0.78 & 0.83 & 0.72 & 0.85 & 0.71 \\[1mm]\hdashline\\[-2mm]
Perturbations (threshold $\approx 3.5$) & 0.66 & 0.75 & \textbf{0.93} & 0.77 & \textbf{0.91} & 0.87 & \textbf{0.91} \\[1mm]\hdashline\\[-2mm]
Sequence labelling (RoBERTa, strict match) & 0.80 & 0.87 & 0.70 & 0.75 & 0.65 & 0.81 & 0.63 \\
Sequence labelling (RoBERTa, partial match) & 0.76 & 0.82 & 0.91 & 0.83  & 0.88 & \textbf{0.88} & 0.88\\[1mm]\hdashline\\[-2mm]
Web search (URL $\cap$) & \textbf{1.00} & \textbf{1.00} & 0.11 & 0.20 & 0.02 & 0.20 & 0.00 \\
Web search (\# of hits) & 0.66 & 0.75 & 0.92 & 0.77 & 0.90 & 0.85 & \textbf{0.91} \\[1mm]
 \midrule
 \multicolumn{8}{c}{\textbf{Text Anonymization Benchmark (test set)}}\\
   \midrule
Majority rule (mask everything) & 0.43 & 0.52 & \textbf{0.96} & 0.60 & 0.93 & \textbf{0.98} & \textbf{0.94}  \\[1mm]\hdashline\\[-2mm]
LLM probabilities (logistic regression) & 0.44 & 0.53 & 0.91 & 0.60 & 0.85 & 0.93 & 0.85 \\
LLM probabilities (AutoML model) & 0.51 & 0.60 & 0.88 & 0.65 & 0.87 & \textbf{0.98} & 0.86 \\[1mm]\hdashline\\[-2mm]
Span classification (AutoML model) & 0.68 & 0.76 & 0.88 & 0.77  & 0.87 & \textbf{0.98} & 0.86 \\[1mm]\hdashline\\[-2mm]
Perturbations (threshold $\approx 10)$ & 0.43 & 0.52 & \textbf{0.96} & 0.60  & \textbf{0.94} & \textbf{0.98} & \textbf{0.94} \\[1mm]\hdashline\\[-2mm]
Sequence labelling (Longformer, strict match) & 0.87 & 0.88 & 0.83 & 0.85  & 0.83 & 0.95 & 0.82 \\
Sequence labelling (Longformer, partial match) & 0.82 & 0.84 & 0.91 & \textbf{0.86}  & 0.90 & \textbf{0.98} & 0.90 \\[1mm]\hdashline\\[-2mm]
Web search (URL $\cap$) & \textbf{0.99} & \textbf{0.99} & 0.12 & 0.22 & 0.03 & 0.43 & 0.01 \\
Web search (\# of hits) & 0.43 & 0.51 & 0.94 & 0.59 & 0.91 & 0.90 & 0.92 \\
\end{tabular}
\end{footnotesize}
\end{center}
\bigskip
\end{table*}

\subsection{Discussion}
\label{sec:discussion}

We now discuss the experimental results of each privacy risk indicator one by one.



\subsubsection*{LLM probabilities}
\label{ssec:ppa}



We can observe that the AutoML model, which consists of an ensemble of supervised classifiers such as decision trees with hyper-parameters optimized on the development set, outperforms a simpler logistic regression model. In other words, the task of predicting high-risk text spans given the aggregated probabilities obtained from a large language model seems to require a non-linear decision boundary. 



\begin{figure}[t]
  \centering
{\includegraphics[width=0.45\textwidth,trim={0 0 17cm 0},clip]{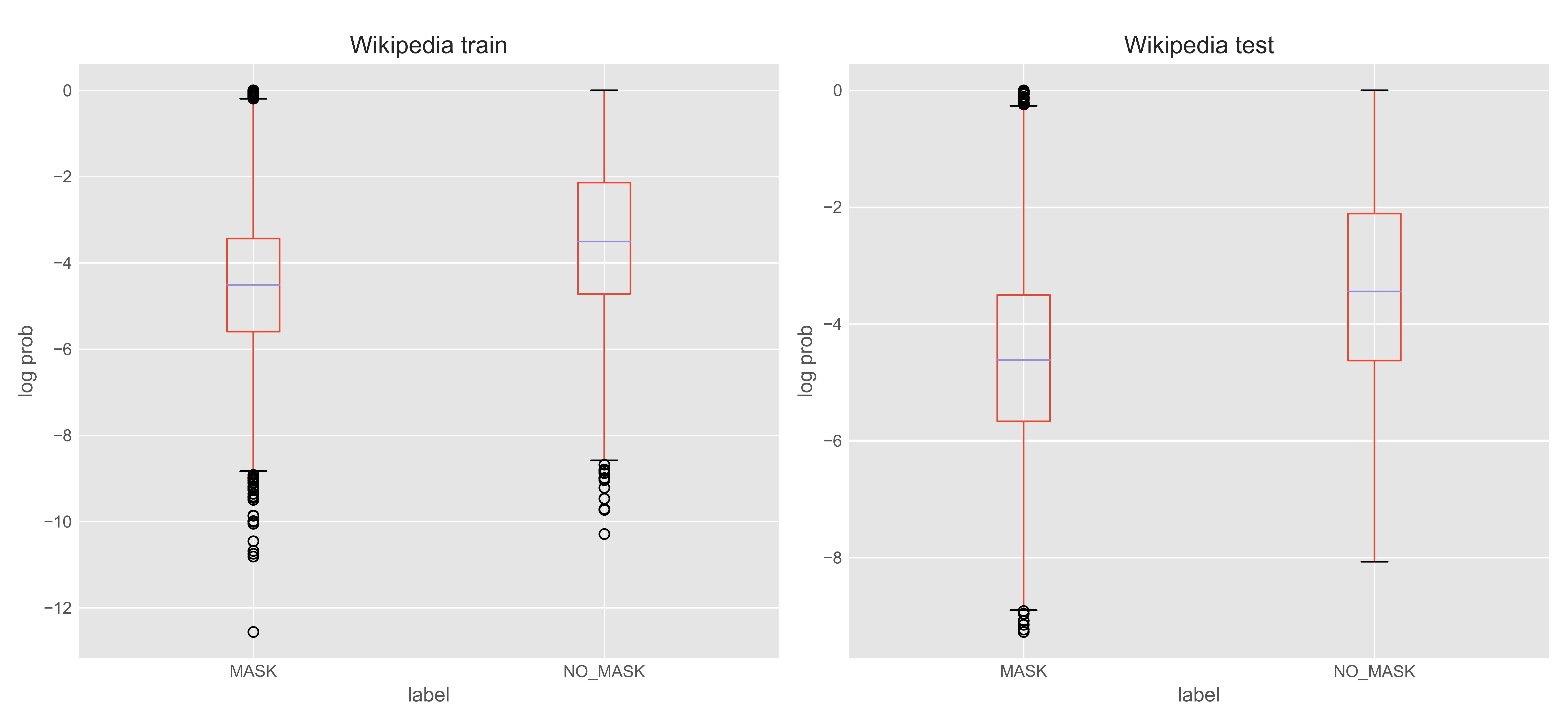}}
  \hfill
{\includegraphics[width=0.45\textwidth,trim={0 0 17cm 0},clip]{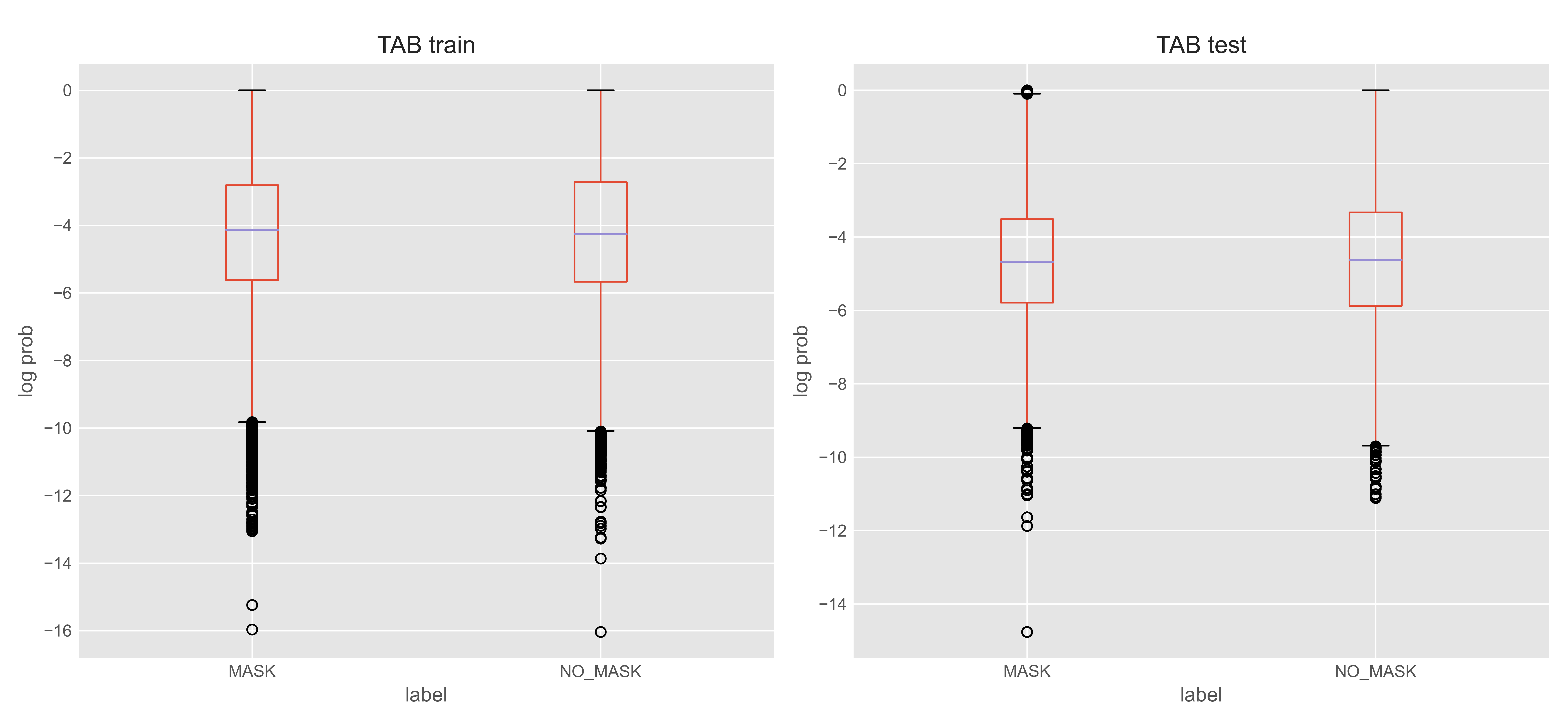}}
  \caption{Token log-probabilities depending on the expert-annotated masking decision (\ENT{MASK} or \ENT{NO\_MASK}) of the corresponding span for Wikipedia and TAB.}
  \label{fig:wiki_tab_graph}
\end{figure}

We also notice a substantial difference between the Wikipedia biographies and the TAB corpus. A closer look at the token log-probabilities factored by masking decision, as shown in Figure \ref{fig:wiki_tab_graph}, is particularly instructive. 
While the PII spans from the Wikipedia biographies show a clear difference between the log-probabilities of the masked and non-masked tokens, this is not the case for the TAB corpus\footnote{At least when aggregating over all PII types. If we do, however, factor those log-probabilities by PII type, we do notice a difference in log-probabilities for a number of PII types.}.

The classifier reaches its optimal performance relatively quickly, after observing about 1 \% of the total training set size in both corpora. Ablation experiments also show that the most important feature for the classifier is the PII type, as some types are masked by human annotators much more often than others. 




\subsubsection*{Span classification}
\label{ssec:scrm}

The span classification approach improves the classification performance compared to the classifier trained only on the aggregated probabilities from the LLM, in particular for the Text Anonymization Benchmark. 

As training this classifier includes fine-tuning a large language model, it requires slighly more training data than the classifier based exclusively on LLM probabilities and PII types. We observe experimentally that the classifier performance stabilizes at around 10\% of training instances for both datasets. Based on ablation experiments, we also see that the outputs of the fine-tuned language model play a major role in the final prediction, along with the PII type.

\subsubsection*{Perturbations}
\label{ssec:threshold}

Despite its theoretical benefits (such as the possibility to directly assess the extent to which a PII span contributes to predicting a direct personal identifier), the perturbation-based method performs poorly, and does not seem to improve upon the majority rule baseline. Indeed, the perturbation mechanism, in combination with the cost function employed to fix the threshold on the log-probabilities, leads to masking the vast majority of text spans.

\subsubsection*{Sequence labelling}
\label{ssec:sc}

Overall, this approach seems to provide the best balance between precision and recall scores, which is to be expected from a large language model fined-tuned on manually labeled data. In contrast to the span classification method, which only considers the tokens inside the PII span itself, the sequence labelling approach can take into account the surrounding context of each span. 

One should note that the results in Tables \ref{tbl:manual} and \ref{tbl:automatic} are obtained with a default threshold of 0.5 on the masking probability. This threshold can, of course, be adjusted to increase the relative importance of recall scores, as is often done in text sanitization to increase the cost of false negatives. 
 
The method based on partial matches, which considers a PII span as risky if at least one constituting token is marked as risky, provides the best results. We also observe that the results on the Wikipedia biographies are lower than those obtained for TAB. This is likely to be due to the smaller number of biographies available for training (453 short texts) compared to the 1014 documents in the training set of TAB.

\subsubsection*{Web Search}
\label{ssec:wq}

Table \ref{tbl:manual} shows a clear difference between using intersection of URLs and using the estimated number of hits, with the first showing a very low recall score while the latter shows a more balanced performance between precision and recall. The intersection of URLs, though, provides better explainability than the number of hits, as one can point the user to the actual URL(s) that contribute to the re-identification. 


While using the intersection of URLs is hampered by the limitation on the number of pages, the chances of a meaningful crossover between target and entities in a text diminishes the deeper into the tail of web results we delve. A possible solution is to try to analyze the text found in the URLs in order to evaluate whether the target individual is mentioned or not.


On the other hand, the number of hits up to a limit is a better approximation of risky entities, despite the unreliability of the number of hits as given by the API. 
Note, though, the restrictions we have already mentioned in Section \ref{ssec:wqra} that are bound to affect the performance. Even though the web can be seen as a very useful and detailed approximation of potential background knowledge an attacker could have and use for re-identification purposes, the technical details of utilizing this make it hard to explain clearly the reason behind the performance. 




\subsubsection*{Summary}
\label{ssec:overall}



As can be expected, privacy risk indicators trained on manually labelled data, in particular sequence labelling, provide the best performance when comparing the spans identified as high risk with expert annotations. Text data annotated with masking decisions is, however, scarce to non-existent for many text sanitization domains. 

When it comes to data utility, we also observe that the precision score weighted by information content ($\mathbf{P}_w$) is higher than the regular precision score for all privacy risk indicators. As argued in \citet{benchmark}, this weighted score is more informative than the basic precision score as it takes into account the informativeness of each token. As a consequence, a higher weighted precision score means that overmasking tends to occur on less informative tokens.

\subsection{Combination of Risk Indicators}
\label{ssec:combos}

Finally, we also evaluate the performance of the privacy risk indicators in combination. More specifically, we consider a manually detected PII span as high-risk if it has been marked as such by \textit{at least} one, two or three risk indicators.\footnote{For the web query risk indicator described in Section \ref{ssec:wqra}, we use the hit-based one since it shows more interesting results. For the sequence labelling model described in Section \ref{ssec:supervised}, we use the spans that were masked under the partial offset match.} The resulting performance is shown in Table \ref{tbl:combinations}.

\begin{table*}[t]
\begin{footnotesize}
\begin{center}
\caption{\label{tbl:combinations}  Evaluation results (on manually labelled PII spans) for the combination of the privacy risk indicators where a PII span is masked if it is considered risky by at least one, two, or three of the five privacy risk indicators presented in the paper.}
\begin{tabular}{*8c}
\toprule
    & \textbf{P} & \textbf{P$_{w}$} & \textbf{ R$_{all}$} & \textbf{F1}  & \textbf{R$_{ent}$} & \textbf{R$_{direct}$} & \textbf{R$_{quasi}$}\\
 \midrule
 \multicolumn{8}{c}{\textbf{Wikipedia biographies (test set)}}\\
 \midrule
   \textit{$\geq$ 1} & 0.78 & 0.82 & \textbf{1.00} & 0.87 & \textbf{1.00} & \textbf{1.00} & \textbf{1.00}\\
  \textit{$\geq$ 2} & 0.80 & 0.84 & \textbf{1.00} & 0.89 & \textbf{1.00} & \textbf{1.00} & \textbf{1.00}\\
  \textit{$\geq$ 3} & \textbf{0.84} & \textbf{0.87} & 0.98 & \textbf{0.90} & 0.96 & \textbf{1.00} & 0.96\\
   \midrule
 \multicolumn{8}{c}{}\\
  \midrule
 \multicolumn{8}{c}{\textbf{Text Anonymization Benchmark (test set)}}\\
 \midrule
\textit{$\geq$ 1} & 0.52 & 0.56 & \textbf{1.00} & 0.68 & \textbf{1.00}& \textbf{1.00} & \textbf{1.00}\\
  \textit{$\geq$ 2} & 0.52 & 0.56 & \textbf{1.00} & 0.69 & \textbf{1.00} & \textbf{1.00} & \textbf{1.00}\\
  \textit{$\geq$ 3} & \textbf{0.61} & \textbf{0.64} & 0.98 & \textbf{0.75} & 0.99 & \textbf{1.00} & 0.99\\
\end{tabular}
\end{center}
\end{footnotesize}
\end{table*}

Although the combination of privacy risk indicators with $\geq 3$ positive signals leads to overmasking, it does detect all direct identifiers and virtually all quasi-identifiers while retaining a higher precision than the majority rule baseline. This high recall is important in text sanitization, as the cost of ignoring a high-risk PII span is much higher than the cost of a false positive. While overmasking does slightly reduce the data utility (by making the text less readable or stripped of some useful content), the presence of a false negative implies that it remains possible to re-identify the individual(s) in question, and that their privacy is therefore not fully ensured.

\section{Conclusions and Future Work}
\label{sec:conclusion}

This paper presented a thorough empirical analysis of two key components of text sanitization methods, namely entity recognition and privacy risk indicators. 


Our experimental results, based on two complementary corpora, indicate that the entity detection task is the most straightforward step to automate. We highlighted the need to go beyond the mere detection of named entities and identify other types of text spans that may provide identifying information. To this end, we showed how to extract large lists of person-related attributes extracted from a knowledge graph such as Wikidata, and thereby apply those lists, along with a standard NER model, to automatically annotate a silver corpus of PII spans.  This silver corpus can, in turn, serve as training data to fine-tune a domain-specific, privacy-oriented entity recognizer.


Determining text spans associated with a high re-identification risk is the most challenging part of the task. A system that masks all detected spans leads to a low privacy risk, but also a comparatively low data utility. 
We presented five distinct indicators of privacy risk, respectively based on LLM probabilities, span classification, perturbations, sequence labelling and web search. Evaluation results comparing the outputs of those approaches with human annotations demonstrate the difficulty of the task. Sequence labelling leads to masking decisions that are most in line with expert annotations, but this approach hinges upon the availability of labelled training data, a requirement that is rarely satisfied in text sanitization tasks. The use of web search also constitutes a promising direction, but suffers from technical constraints arising from the reliance on search engine APIs.


The present paper did not elaborate on how masking should be performed. A common option is to simply replace the selected spans by a black box or a generic string such as ``***''. However, to mitigate the loss of data utility , it is often preferable to replace those PII spans with other, less specific spans, such as editing ``Oslo'' into ``[city in Norway]'' or ``January 6, 2022'' into ``2022''. \citet{olstad-etal-2023-generation} explore an approach for selecting suitable replacements for different types of PII spans, which we evaluated against human decisions. Such an approach could be extended to different domains and different types of PII spans. 

Evaluating text sanitization tasks is a challenging problem, notably due to the presence of multiple equally correct solutions for a given document \citep{lison-etal-2021-anonymisation}. In this work we evaluate against manually annotated documents, using several expert annotators for each document to ensure a reasonable set of possible solutions. Relying on manual annotations is, however, not always feasible, so alternative evaluation methods suitable for the task should also be explored. \textit{Re-identification} constitutes a promising approach \citep{scaiano2016unified,mozes2021no,10.1007/978-3-031-13945-1_12}, and operates by carrying out an attack aiming to determine whether an individual was part of the sanitized collection of documents, often having access to additional, external resources. Future work will look at the use of such re-identification attacks as an alternative measure of the text sanitization strength.


\section*{Declarations}

\subsection*{Author Contribution}
A.P: conceptualization, implementation of privacy risk indicators (apart from web search), evaluation of privacy-aware entity recognition, evaluation of privacy risk indicators alone and in combination, creation of figures, creation of table, writing and editing of the final manuscript. P.L: supervision, writing and editing of the final manuscript. M.A: web search risk indicator, writing. L.Ø: supervision, writing. I.P: supervision, writing. All authors reviewed the manuscript prior to submission.

\subsection*{Funding}
The research leading to these results received funding from the Research Council of Norway (CLEANUP project) under Grant nr. 308904.

\subsection*{Conflict of Interest}
The authors have no conflict of interest to declare.

\subsection*{Data Availability Statement}
No datasets were generated during the current study.

\subsection*{Acknowledgment}

We acknowledge support from the Norwegian Research Council (CLEANUP project, grant nr.308904).

\newpage

\appendix
\appendixpage

\section{Human properties from Wikidata}
\label{human_properties}

The two tables below show the selected Wikidata properties mentioned in Section \ref{sec:properties} that constitute the \ENT{DEM} and \ENT{MISC} gazetteers.

\subsection*{DEM-related properties}

\begin{longtblr}{|p{28mm}|p{28mm}|p{63mm}|} \hline
        \textbf{Property} & \textbf{Example} & \textbf{Description} \\ \hline
        \raggedright hairstyle / hairlength  (P8839) & \raggedright   afro & style of cutting, arranging, or combing the hair on the subject's scalp \\ \hline
        \raggedright grants  (P5460) & \raggedright   Bachelor of Arts & confers degree, honor, award, prize, title, certificate or medal denoting achievement to a person or organization \\ \hline
        \raggedright next lower rank  (P3729) & \raggedright   General & ``lower rank or level in a ranked hierarchy like sport league, military ranks. If there are several possible, list each one and qualify with ````criterion used"" (P1013), avoid using ranks and date qualifiers. For sports leagues/taxa, use specific properties instead." \\ \hline
        \raggedright special rank  (P5012) & \raggedright   General of the Police of the Russian Federation & non-military rank given to a civil office holder in Russia \\ \hline
        \raggedright political ideology  (P1142) & \raggedright   social liberalism & political ideology of an organization or person or of a work (such as a newspaper) \\ \hline
        \raggedright office held by head of the organization  (P2388) & \raggedright   Chief Justice of the United States & position of the head of this item \\ \hline
        \raggedright ethnic group  (P172) & \raggedright   Ashkenazi Jews & subject's ethnicity (consensus is that a VERY high standard of proof is needed for this field to be used. In general this means 1) the subject claims it themselves, or 2) it is widely agreed on by scholars, or 3) is fictional and portrayed as such) \\ \hline
        \raggedright facial hair  (P8852) & \raggedright   chin curtain & style of cutting, arranging, or combing the hair on the subject's face (usually on the chin, cheeks, and upper lip region); beard \\ \hline
        \raggedright language used  (P2936) & \raggedright   Catalan & language widely used (spoken or written) in this place or at this event \\ \hline
        \raggedright writing language  (P6886) & \raggedright   French & language in which the writer has written their work \\ \hline
        \raggedright religion  (P8929) & \raggedright   agnosticism & religious or world view affiliation of a person, organization or religious building \\ \hline
        \raggedright sex or gender  (P21) & \raggedright   male organism & sex or gender identity of human or animal. For human: male, female, non-binary, intersex, transgender female, transgender male, agender. For animal: male organism, female organism. Groups of same gender use subclass of (P279) \\ \hline
        \raggedright languages spoken, written or signed  (P1412) & \raggedright   Old Chinese & language(s) that a person or a people speaks, writes or signs, including the native language(s) \\ \hline
        \raggedright noble title  (P97) & \raggedright   baron & titles held by the person \\ \hline
        \raggedright position held  (P39) & \raggedright   President of the United States & subject currently or formerly holds the object position or public office \\ \hline
        \raggedright Y-DNA Haplogroup  (P4426) & \raggedright   Haplogroup R1b & Y-DNA haplogroup of a person or organism \\ \hline
        \raggedright original language of film or TV show  (P364) & \raggedright   italian & ``language in which a film or a performance work was originally created. Deprecated for written works and songs; use P407 (````language of work or name"") instead." \\ \hline
        \raggedright canonization status  (P411) & \raggedright   saint & stage in the process of attaining sainthood per the subject's religious organization \\ \hline
        \raggedright occupation  (P106) & \raggedright   writer & ``occupation of a person; see also ````field of work"" (Property:P101), ````position held"" (Property:P39)" \\ \hline
        \raggedright eye color  (P1340) & \raggedright   blue-green & color of the irises of a person's eyes \\ \hline
        \raggedright culture  (P2596) & \raggedright   Inca Empire & human culture or people (or several cultures) associated with this item \\ \hline
        \raggedright practiced by  (P3095) & \raggedright   cheerleader & type of agents that study this subject or work in this profession \\ \hline
        \raggedright official language  (P37) & \raggedright   English & language designated as official by this item \\ \hline
        \raggedright dialect of  (P4913) & \raggedright   Northern Sami & ``language of which an item with this property is a dialect. Use in addition to ````subclass of"" (P279) if a languoid is also considered a dialect." \\ \hline
        \raggedright political alignment  (P1387) & \raggedright   left-wing & political position within the left-right political spectrum \\ \hline
        \raggedright educational stage  (P7374) & \raggedright   undergraduate education & ``educational stage of a specific school; qualifier of ````educated at"" (P69): the education stage of someone's attendance of school" \\ \hline
        \raggedright substitute  (P2098) & \raggedright   Lieutenant Governor of the United States Virgin Islands & function that serves as deputy/replacement of this function/office (scope/conditions vary depending on office) \\ \hline
        \raggedright professorship  (P803) & \raggedright   Lucasian Professor of Mathematics & professorship position held by this academic person \\ \hline
        \raggedright official religion  (P3075) & \raggedright   Islam & official religion in this administrative entity \\ \hline
        \raggedright office contested  (P541) & \raggedright   Chief Executive of Hong Kong & title of office which election will determine the next holder of \\ \hline
        \raggedright sexual orientation  (P91) & \raggedright   homosexuality & the sexual orientation of the person -- use IF AND ONLY IF they have stated it themselves, unambiguously, or it has been widely agreed upon by historians after their death \\ \hline
        \raggedright kinship to subject  (P1039) & \raggedright   daughter-in-law & ``qualifier of ````relative"" (P1038) to indicate less usual family relationships (ancestor, son-in-law, adoptions, etc). Indicate how the qualificator item is related to the main item." \\ \hline
        \raggedright next higher rank  (P3730) & \raggedright   pope & ``higher rank or level in a ranked hierarchy like sport league, military ranks. If there are several possible, list each one and qualify with ````criterion used"" (P1013), avoid using ranks and date qualifiers. For sports leagues/taxa, use specific properties instead." \\ \hline
        \raggedright academic degree  (P512) & \raggedright   Doctor of Philosophy & academic degree that the person holds \\ \hline
        \raggedright religion or worldview  (P140) & \raggedright   Hinduism & religion of a person, organization or religious building, or associated with this subject \\ \hline
        \raggedright office held by head of government  (P1313) & \raggedright   Prime Minister of the United Kingdom & political office that is fulfilled by the head of the government of this item \\ \hline
        \raggedright native language  (P103) & \raggedright   Russian & language or languages a person has learned from early childhood \\ \hline
        \raggedright members have occupation  (P3989) & \raggedright   lawyer & all members of this group share the occupation \\ \hline
        \raggedright language of work or name  (P407) & \raggedright   Italian & ``language associated with this creative work (such as books, shows, songs, or websites) or a name (for persons use ````native language"" (P103) and ````languages spoken, written or signed"" (P1412))" \\ \hline
        \raggedright office held by head of state  (P1906) & \raggedright   President of Israel & political office that is fulfilled by the head of state of this item \\ \hline
        \raggedright military rank  (P410) & \raggedright   General & ``military rank achieved by a person (should usually have a ````start time"" qualifier), or military rank associated with a position" \\ \hline
        \raggedright medical condition  (P1050) & \raggedright   spina bifida & any state relevant to the health of an organism, including diseases and positive conditions \\ \hline
        \raggedright hair color  (P1884) & \raggedright   brown hair & person's hair color. Use P585 as qualifier if there's more than one value \\ \hline
        \raggedright title of chess person  (P2962) & \raggedright   Grandmaster & title awarded by a chess federation to a person \\ \hline
    \end{longtblr}

    \subsection*{MISC-related properties}

\begin{longtblr}{|p{28mm}|p{28mm}|p{63mm}|} \hline
    \textbf{Property} & \textbf{Example} & \textbf{Description} \\ \hline
        \raggedright classification of race (P2502) & \raggedright 2015 Schaal Sels & race for which this classification applies \\ \hline
        \raggedright contributing factor of (P1537) & \raggedright critical thinking & thing that is significantly influenced by this cause, but does not directly result from it.  See 'Help:Modeling causes' for examples and discussion. \\ \hline
        \raggedright length (P2043) & \raggedright 15 centimeters & measured dimension of an object \\ \hline
        \raggedright conflict (P607) & \raggedright Battle of Dennewitz & battles, wars or other military engagements in which the person or item participated \\ \hline
        \raggedright school of (P1780) & \raggedright Ain AlBaraha & for a creative work with author with a style influenced by the known author or circle, active in the same period, but a student or follower \\ \hline
        \raggedright bowling style (P2545) & \raggedright Left-arm orthodox spin & type of bowling employed by a cricketer \\ \hline
        \raggedright field of training (P8258) & \raggedright bookselling & profession, position or activity someone learnt during a training, traineeship or apprenticeship \\ \hline
        \raggedright used by (P1535) & \raggedright television watcher & item or concept that makes use of the subject (use sub-properties when appropriate) \\ \hline
        \raggedright studied by (P2579) & \raggedright egyptology & subject is studied by this science or domain \\ \hline
        \raggedright hashtag (P2572) & \raggedright BlackLivesMatter & ``hashtag associated with this item. Format: do not include the ````\#"" symbol" \\ \hline
        \raggedright worshipped by (P1049) & \raggedright Zoroastrianism & religion or group/civilization that worships a given deity \\ \hline
        \raggedright motto (P1546) & \raggedright Florebo quocumque ferar & description of the motto of the subject \\ \hline
        \raggedright statement is subject of (P805) & \raggedright death of Osama bin Laden & (qualifying) item that describes the relation identified in this statement \\ \hline
        \raggedright GitHub topic (P9100) & \raggedright bitcoin & GitHub topic for a repository or subject \\ \hline
        \raggedright original spelling (P1353) & \raggedright Aloe steffaniana & original spelling of a scientific name \\ \hline
        \raggedright armament (P520) & \raggedright AIM\-120 AMRAAM & equippable weapon item for the subject \\ \hline
        \raggedright architectural style (P149) & \raggedright French Gothic architecture & architectural style of a structure \\ \hline
        \raggedright academic major (P812) & \raggedright political science & major someone studied at college/university \\ \hline
        \raggedright in opposition to (P5004) & \raggedright Plan Juppé, 1995 & qualifier to indicate the purpose of a social action \\ \hline
        \raggedright astronaut mission (P450) & \raggedright Soyuz TMA-8 & space mission that the subject is or has been a member of (do not include future missions) \\ \hline
        \raggedright field of this occupation (P425) & \raggedright bodyboarding & activity corresponding to this occupation or profession (use only for occupations/professions - for people use Property:P101, for companies use P452) \\ \hline
        \raggedright vessel (P1876) & \raggedright Soyuz-TMA & vessel involved in this mission, voyage or event \\ \hline
        \raggedright test taken (P5021) & \raggedright SAT & subject took the test or exam, or was assessed per test or method \\ \hline
        \raggedright trophy awarded (P4622) & \raggedright Oscar statuette & trophy awarded at the end of a selection process or of a competition, usually to winners or finishers under the form of a cup, a medal, a diploma \\ \hline
        \raggedright sport number (P1618) & \raggedright 27 & number worn on a player's (competitor's) uniform, equipment, etc \\ \hline
        \raggedright Twitter username (P2002) & \raggedright NASA & this item's username on Twitter; do not include the ??? symbol \\ \hline
        \raggedright programming language (P277) & \raggedright C++ & the programming language(s) in which the software is developed \\ \hline
        \raggedright advertises (P6291) & \raggedright Michelin & cause, organization or product targeted by the advertising campaign, character, jingle or slogan \\ \hline
        \raggedright uses (P2283) & \raggedright paintbrush & item or concept used by the subject or in the operation (see also instrument [P1303] and armament [P520]) \\ \hline
        \raggedright study or design for (P6606) & \raggedright Birds and Irises & finished work of art for which the subject is a preliminary study, design, or conceptual drawing \\ \hline
        \raggedright manner of death (P1196) & \raggedright capital punishment & general circumstances of a person's death; e.g. natural causes, accident, suicide, homicide, etc.  Use 'cause of death' (P509) for the specific physiological mechanism, e.g. heart attack, trauma, pneumonia... \\ \hline
        \raggedright evaluation of (P5134) & \raggedright heart rate & the subject finding is an evaluation or interpretation of the object attribute or test \\ \hline
        \raggedright Eight Banner register (P470) & \raggedright Manchu Bordered Yellow Banner & Manchu household register for people of the Qing Dynasty \\ \hline
        \raggedright title of broader work (P6333) & \raggedright Revue des Deux Mondes & ``title of website, journal or other work in which an article (or other publication) is published in. Use ````published in"" (P1433) instead if possible." \\ \hline
        \raggedright genre (P136) & \raggedright war film & creative work's genre or an artist's field of work (P101). Use main subject (P921) to relate creative works to their topic \\ \hline
        \raggedright applies to jurisdiction (P1001) & \raggedright Eurozone & the item (institution, law, public office, public register...) or statement belongs to or has power over or applies to the value (a territorial jurisdiction: a country, state, municipality, ...) \\ \hline
        \raggedright playing hand (P741) & \raggedright left-handedness & hand used to play a racket sport \\ \hline
        \raggedright fictional analog of (P1074) & \raggedright spacecraft & used to link an entity or class of entities appearing in a creative work with the analogous entity or class of entities in the real world \\ \hline
        \raggedright inscription (P1684) & \raggedright O U O S V A V V & inscriptions, markings and signatures on an object \\ \hline
        \raggedright broader concept (P4900) & \raggedright hat & qualifier to indicate a broader concept that the present item is part of, as mapped by an external source. The statement being qualified should be an exact match. \\ \hline
        \raggedright legal citation of this text (P1031) & \raggedright 384 U.S. 436 & legal citation of legislation or a court decision \\ \hline
        \raggedright exhibition history (P608) & \raggedright Saint Anne, Leonardo da Vinci ultimate masterpiece & exhibitions where the item is or was displayed \\ \hline
        \raggedright form of creative work (P7937) & \raggedright novel & structure of a creative work, type of music release \\ \hline
        \raggedright has edition or translation (P747) & \raggedright Orgoglio e pregiudizio & link to an edition of this item \\ \hline
        \raggedright nominated for (P1411) & \raggedright BRIT Awards & ``award nomination received by a person, organisation or creative work (inspired from ````award received"" (Property:P166))" \\ \hline
        \raggedright voice type (P412) & \raggedright lyric spinto tenor & person's voice type. expected values: soprano, mezzo-soprano, contralto, countertenor, tenor, baritone, bass (and derivatives) \\ \hline
        \raggedright platform (P400) & \raggedright Xbox 360 & platform for which a work was developed or released, or the specific platform version of a software product \\ \hline
        \raggedright drug used for treatment (P2176) & \raggedright dabrafenib & drug, procedure, or therapy that can be used to treat a medical condition \\ \hline
        \raggedright owner of (P1830) & \raggedright SPY Records & entities owned by the subject \\ \hline
        \raggedright fabrication method (P2079) & \raggedright sericulture & method, process or technique used to grow, cook, weave, build, assemble, manufacture the item \\ \hline
        \raggedright connects with (P2789) & \raggedright Rokin & item with which the item is physically connected \\ \hline
        \raggedright totem (P2831) & \raggedright Siluriformes & in many indigenous cultures an individual or group has a particular totem (e.g. a type of animal) \\ \hline
        \raggedright field of work (P101) & \raggedright linguistic purism & specialization of a person or organization; see P106 for the occupation \\ \hline
        \raggedright definition domain (P1568) & \raggedright set of real numbers & ``set of ````input"" or argument values for which a mathematical function is defined" \\ \hline
        \raggedright parliamentary term (P2937) & \raggedright 13th legislature of the Fifth French Republic & term of a parliament or any deliberative assembly \\ \hline
        \raggedright exonerated of (P7781) & \raggedright murder & crime of which the subject was found not to be guilty after having previously been found guilty by a court for that crime \\ \hline
        \raggedright carries scientific instrument (P1202) & \raggedright Astro & scientific instruments carried by a vessel, satellite, or device that are not required for propelling or navigating \\ \hline
        \raggedright distribution format (P437) & \raggedright Blu-ray Disc & method (or type) of distribution for the subject \\ \hline
        \raggedright use (P366) & \raggedright source of information & main use of the subject (includes current and former usage) \\ \hline
        \raggedright end cause (P1534) & \raggedright resignation & qualifier to use together with the end date qualifier (P582) to specify the cause \\ \hline
        \raggedright source of income (P2770) & \raggedright donation & source of income of an organization or person \\ \hline
        \raggedright participant in (P1344) & \raggedright 1992 Winter Olympics & event in which a person or organization was/is a participant; inverse of P710 or P1923 \\ \hline
        \raggedright depicted by (P1299) & \raggedright Traveler Over The Mist Sea & object depicting this subject (creative works, books of the bible) \\ \hline
        \raggedright member of the crew of (P5096) & \raggedright Titanic & person who has been a member of a crew associated with the vessel or spacecraft. For spacecraft, inverse of crew member (P1029), backup or reserve team or crew (P3015) \\ \hline
        \raggedright victory (P2522) & \raggedright 2015 Internationale Wielertrofee Jong Maar Moedig & competition or event won by the subject \\ \hline
        \raggedright elected in (P2715) & \raggedright 1999 German presidential election & qualifier for statements in position held to link the election in which a person gained a position from, or reelection in which the position was confirmed \\ \hline
        \raggedright has effect (P1542) & \raggedright mercury poisoning & effect of this item \\ \hline
        \raggedright corrigendum / erratum (P2507) & \raggedright Evolution in Mendelian Populations & published notice that corrects or amends a previous publication \\ \hline
        \raggedright electoral district (P768) & \raggedright Ohio's 8th congressional district & ``electoral district this person is representing, or of the office that is being contested. Use as qualifier for ````position held"" (P39) or ````office contested"" (P541) or ````candidacy in election"" (P3602)" \\ \hline
        \raggedright mascot (P822) & \raggedright Scotty the Scottie Dog & mascot of an organization, e.g. a sports team or university \\ \hline
        \raggedright facet of (P1269) & \raggedright general relativity & topic of which this item is an aspect, item that offers a broader perspective on the same topic \\ \hline
        \raggedright has cause (P828) & \raggedright mercury & underlying cause, thing that ultimately resulted in this effect \\ \hline
        \raggedright direction (P560) & \raggedright north & qualifier to indicate the direction of the object relative to the subject item (for direction to the object, see P654) \\ \hline
        \raggedright part of (P361) & \raggedright head & ``object of which the subject is a part (if this subject is already part of object A which is a part of object B, then please only make the subject part of object A). Inverse property of ````has part"" (P527, see also ````has parts of the class"" (P2670))." \\ \hline
        \raggedright lifestyle (P1576) & \raggedright vegetarianism & typical way of life of an individual, group, or culture \\ \hline
        \raggedright writing system (P282) & \raggedright English alphabet & alphabet, character set or other system of writing used by a language, supported by a typeface \\ \hline
        \raggedright follows (P155) & \raggedright Star Wars: Episode IV A New Hope & ``immediately prior item in a series of which the subject is a part, preferably use as qualifier of P179 [if the subject has replaced the preceding item, e.g. political offices, use ````replaces"" (P1365)]" \\ \hline
        \raggedright flag (P163) & \raggedright flag of Slovakia & subject's flag \\ \hline
        \raggedright part of the series (P179) & \raggedright The Chronicles of Narnia & series which contains the subject \\ \hline
        \raggedright means of locomotion (P3512) & \raggedright bipedalism & method that the subject uses to move from one place to another \\ \hline
        \raggedright color (P462) & \raggedright black & color of subject \\ \hline
        \raggedright constellation (P59) & \raggedright Andromeda & the area of the celestial sphere of which the subject is a part (from a scientific standpoint, not an astrological one) \\ \hline
        \raggedright medical examinations (P923) & \raggedright medical radiography & examinations that might be used to diagnose and/or prognose the medical condition \\ \hline
        \raggedright partially coincident with (P1382) & \raggedright Interstate 76 & object that is partially part of, but not fully part of (P361), the subject \\ \hline
        \raggedright catchphrase (P6251) & \raggedright Did I do that? & commonly used phrase by which someone or something may be recognized \\ \hline
        \raggedright stock exchange (P414) & \raggedright New York Stock Exchange & exchange on which this company is traded \\ \hline
        \raggedright motto text (P1451) & \raggedright Un Dieu, un roy, une foy, une loy (French) & short motivation sentence associated to item \\ \hline
        \raggedright level of description (P6224) & \raggedright record group & position of the unit of description in the hierarchy of the fonds \\ \hline
        \raggedright instrumentation (P870) & \raggedright piano & combination of musical instruments employed in a composition or accompanying a (folk) dance \\ \hline
        \raggedright footedness (P8006) & \raggedright left-footedness & dominant foot or preferred stance of this person \\ \hline
        \raggedright subclass of (P279) & \raggedright fruit & next higher class or type; all instances of these items are instances of those items; this item is a class (subset) of that item. Not to be confused with P31 (instance of) \\ \hline
        \raggedright domain of saint or deity (P2925) & \raggedright sea & domain(s) which this saint or deity controls or protects \\ \hline
        \raggedright charted in (P2291) & \raggedright Billboard 200 & chart where the element reached a position \\ \hline
        \raggedright radio format (P415) & \raggedright rhythmic contemporary & describes the overall content broadcast on a radio station \\ \hline
        \raggedright category contains (P4224) & \raggedright architectural structure & category contains elements that are instances of this item \\ \hline
        \raggedright theme music (P942) & \raggedright St. Anger & the theme music/song used by the item \\ \hline
        \raggedright sport (P641) & \raggedright trail running & sport that the subject participates or participated in or is associated with \\ \hline
        \raggedright mount (P3091) & \raggedright Rocinante & creature ridden by the subject, for instance a horse \\ \hline
        \raggedright notable work (P800) & \raggedright Hamlet & notable scientific, artistic or literary work, or other work of significance among subject's works \\ \hline
        \raggedright movement (P135) & \raggedright Frankfurt School & literary, artistic, scientific or philosophical movement or scene associated with this person or work \\ \hline
        \raggedright has pattern (P5422) & \raggedright stripe & pattern, design, or motif intrinsic to, incorporated into, or applied to the surface of the subject \\ \hline
        \raggedright history of topic (P2184) & \raggedright history of Germany & ``item about the historical development of an subject's topic, sample: ````history of Argentina"" for ````Argentina"". To list key events of the topic, use ````significant event"" (P793)" \\ \hline
        \raggedright position played on team / speciality (P413) & \raggedright centre & position or specialism of a player on a team \\ \hline
        \raggedright basic form of government (P122) & \raggedright unitary state & subject's government \\ \hline
        \raggedright call sign (P2317) & \raggedright Freedom 7 & short form identifier for a radio operator or broadcaster and (as flag signal) previous for watercrafts \\ \hline
        \raggedright statistical unit (P2353) & \raggedright country & member of a dataset \\ \hline
        \raggedright cabinet (P5054) & \raggedright Florida Cabinet & ``qualifier for ````position held"" (P39) to indicate the cabinet to which belongs a minister" \\ \hline
        \raggedright main subject (P921) & \raggedright Rocco Granata & primary topic of a work (see also P180: depicts) \\ \hline
        \raggedright Artsy gene (P2411) & \raggedright cubism & generalization of artwork type, technique, material, genre, movement, etc. from artsy.net \\ \hline
        \raggedright template has topic (P1423) & \raggedright Syngnathiformes & topic related to template \\ \hline
        \raggedright candidacy in election (P3602) & \raggedright 2012 Finnish municipal elections & election where the subject is a candidate \\ \hline
        \raggedright does not have part (P3113) & \raggedright control tower & expected part that the item does not have (for qualities, use P6477) \\ \hline
        \raggedright commemorates (P547) & \raggedright Johann Sebastian Bach & what the place, monument, memorial, or holiday, commemorates \\ \hline
        \raggedright alternate names (P4970) & \raggedright hankies & qualifier for alternative name(s) given for a subject in a database entry \\ \hline
        \raggedright iconographic symbol (P4185) & \raggedright Keys of Heaven & identifying element typically depicted as accompanying or worn by this religious figure, hero, fictional or historical character \\ \hline
        \raggedright cuisine (P2012) & \raggedright Indian cuisine & type of food served by a restaurant or restaurant chain or national food culture \\ \hline
        \raggedright possible medical findings (P5131) & \raggedright mid-systolic click & possible medical findings of a medical condition \\ \hline
        \raggedright parent taxon (P171) & \raggedright Sus scrofa & closest parent taxon of the taxon in question \\ \hline
        \raggedright presented in (P5072) & \raggedright TED2010 & event at which a talk, film, paper, or poster was presented \\ \hline
        \raggedright title (P1476) & \raggedright Nature & published name of a work, such as a newspaper article, a literary work, piece of music, a website, or a performance work \\ \hline
        \raggedright affiliation string (P6424) & \raggedright Southampton University & qualifier to provide the published string form of affiliation attached to an author \\ \hline
        \raggedright quotation or excerpt (P7081) & \raggedright who would these fardels bear[...] & quotation or excerpt from this work. No quotation marks needed \\ \hline
        \raggedright subject has role (P2868) & \raggedright antineoplastic & ``role/generic identity of the item (````subject""), also in the context of a statement. For the role of the value of the statement (````object""), use P3831 (````object has role""). For acting roles, use P453 (````character role""). For persons, use P39." \\ \hline
        \raggedright item operated (P121) & \raggedright transport infrastructure & equipment, installation or service operated by the subject \\ \hline
        \raggedright sports competition competed at (P5249) & \raggedright 2012 UEFA Champions League & edition of sports competitions at which the club or team that played a season competed at. Use this to link items for team seasons to items for specific seasons of competitions.Sample: 2012 ??? 13 FC Barcelona season ??? 13 UEFA Champions League \\ \hline
        \raggedright literal translation (P2441) & \raggedright first son (English) & direct or word-for-word translation of a name or phrase (qualifier for name, title, inscription, and quotation properties) \\ \hline
        \raggedright from narrative universe (P1080) & \raggedright Tolkien's legendarium & subject's fictional entity is in the object narrative. See also P1441 (present in work) and P1445 (fictional universe described in) \\ \hline
        \raggedright seal description (P418) & \raggedright Great Seal of the United States & links to the item for the subject's seal \\ \hline
        \raggedright anthem (P85) & \raggedright State Anthem of the Soviet Union & subject's official anthem \\ \hline
        \raggedright taxon name (P225) & \raggedright Panthera leo & correct scientific name of a taxon (according to the reference given) \\ \hline
        \raggedright wears (P3828) & \raggedright engagement announcement dress of Catherine Middleton & clothing or accessory worn on subject's body \\ \hline
        \raggedright studies (P2578) & \raggedright Ancient Egypt & the object that an academic field studies; distinct from field of work (P101), which is used for human, organization, etc. \\ \hline
        \raggedright business model (P7936) & \raggedright free-to-play & business model a business enterprise or creative work operates under \\ \hline
        \raggedright social classification (P3716) & \raggedright commoner & social class as recognized in traditional or state law \\ \hline
        \raggedright record held (P1000) & \raggedright men's pole vault world record & notable record achieved by a person or entity, include qualifiers for dates held \\ \hline
        \raggedright named after (P138) & \raggedright Camembert & ``entity or event that inspired the subject's name, or namesake (in at least one language). Qualifier ````applies to name"" (P5168) can be used to indicate which one" \\ \hline
        \raggedright general classification of race participants (P2321) & \raggedright Nacer Bouhanni & classification of race participants \\ \hline
        \raggedright tracklist (P658) & \raggedright Wanna Be Startin' Somethin' & audio tracks contained in this release \\ \hline
        \raggedright nickname (P1449) & \raggedright The Golden State & informal name (for a pseudonym use P742) \\ \hline
        \raggedright shooting handedness (P423) & \raggedright left-handed shot & whether the hockey player passes or shoots left- or right-handed \\ \hline
        \raggedright significant event (P793) & \raggedright Battle of Dennewitz & significant or notable events associated with the subject \\ \hline
        \raggedright followed by (P156) & \raggedright April & ``immediately following item in a series of which the subject is a part, preferably use as qualifier of P179 [if the subject has been replaced, e.g. political offices, use ````replaced by"" (P1366)]" \\ \hline
        \raggedright manner of inhumane treatment (P7160) & \raggedright sleep deprivation & manner of torture and other cruel, inhumane or degrading treatment or punishments as covered by the UN Convention experienced by the subject \\ \hline
        \raggedright possessed by spirit (P4292) & \raggedright Pazuzu & item which is spiritually possessing this item \\ \hline
        \raggedright fandom (P8693) & \raggedright A.I Troops & fan group name of a celebrity, musical group, or an artist \\ \hline
        \raggedright input method (P479) & \raggedright computer keyboard & input method or device used to interact with a software product \\ \hline
        \raggedright damaged (P3081) & \raggedright Upper Basilica of San Francesco d'Assisi & physical items damaged by this event \\ \hline
        \raggedright plea (P1437) & \raggedright guilt & whether a person pleaded guilty, not guilty, etc. \\ \hline
        \raggedright derivative work (P4969) & \raggedright MX Linux & new work of art (film, book, software, etc.) derived from major part of this work \\ \hline
        \raggedright dataset distribution (P2702) & \raggedright Art \& Architecture Thesaurus LOD Dataset & particular manner of distribution of a data set (database or file) that is publicly available \\ \hline
        \raggedright real estate developer (P6237) & \raggedright Strabag Real Estate & person or organization responsible for building this item \\ \hline
        \raggedright product or material produced (P1056) & \raggedright uranium & material or product produced by a government agency, business, industry, facility, or process \\ \hline
        \raggedright contributed to creative work (P3919) & \raggedright Dictionary of National Biography & ``person is cited as contributing to some creative or published work or series (qualify with ````subject has role"", P2868)" \\ \hline
        \raggedright military casualty classification (P1347) & \raggedright killed in action & allowed values:  killed in action (Q210392), missing in action (Q2344557), died of wounds (Q16861372), prisoner of war (Q179637), killed in flight accident (Q16861407), others used in military casualty classification \\ \hline
        \raggedright separated from (P807) & \raggedright Presbyterian Church in the United States & subject was founded or started by separating from identified object \\ \hline
        \raggedright results (P2501) & \raggedright general classification of the 2015 Coupe Sels & results of a competition such as sports or elections \\ \hline
        \raggedright ritual object (P8706) & \raggedright corpse powder & ceremonial or ritual objects associated with, and used as part of, rites and rituals practiced in everyday life and in rarer cultic and communal rites \\ \hline
        \raggedright edition or translation of (P629) & \raggedright Pride and Prejudice & is an edition or translation of this entity \\ \hline
        \raggedright Catholic rite (P3501) & \raggedright Roman Rite & Christian liturgical rite associated with this item \\ \hline
        \raggedright academic minor (P811) & \raggedright political science & minor someone studied at college/university \\ \hline
        \raggedright intangible cultural heritage status (P3259) & \raggedright inventory of intangible cultural heritage in France & status of an item that is designated as intangible heritage \\ \hline
        \raggedright opposite of (P461) & \raggedright peace & item that is the opposite of this item \\ \hline
        \raggedright reward (P4444) & \raggedright travel insurance & reward, bonus, or prize received as part of a membership or loyalty program \\ \hline
        \raggedright award received (P166) & \raggedright Theodor Haecker Price & award or recognition received by a person, organisation or creative work \\ \hline
        \raggedright legal form (P1454) & \raggedright GmbH \& Co. KG & legal form of an entity \\ \hline
        \raggedright Bharati Braille (P9021) & \raggedright braille script & transcription of Indic scripts in a national standard Braille script \\ \hline
        \raggedright vehicle normally used (P3438) & \raggedright popemobile & vehicle the subject normally uses \\ \hline
        \raggedright coat of arms (P237) & \raggedright Coat of arms of Liechtenstein & subject's coat of arms \\ \hline
        \raggedright convicted of (P1399) & \raggedright murder & crime a person was convicted of \\ \hline
        \raggedright made from material (P186) & \raggedright chocolate liquor & material the subject or the object is made of or derived from \\ \hline
        \raggedright competition class (P2094) & \raggedright Group R & official classification by a regulating body under which the subject (events, teams, participants, or equipment) qualifies for inclusion \\ \hline
        \raggedright shape (P1419) & \raggedright pyramid & shape of an object \\ \hline
        \raggedright academic thesis (P1026) & \raggedright Pan, Sinking: for Steelpan and Nine Instruments & thesis or dissertation written for a degree \\ \hline
        \raggedright audio system (P7501) & \raggedright bell & audio system hardware used in the item \\ \hline
        \raggedright first appearance (P4584) & \raggedright A Study in Scarlet & work in which a fictional/mythical character or entity first appeared \\ \hline
        \raggedright last words (P3909) & \raggedright I want nothing but death. & last words attributed to a person before their death \\ \hline
        \raggedright strand orientation (P2548) & \raggedright software & orientation of gene on double stranded DNA molecule \\ \hline
        \raggedright female form of label (P2521) & \raggedright autora & female form of name or title \\ \hline
        \raggedright penalty (P1596) & \raggedright capital punishment & penalty imposed by an authority \\ \hline
        \raggedright handedness (P552) & \raggedright right-handedness & handedness of the person \\ \hline
        \raggedright Google Play developer slug (P8939) & \raggedright Facebook & human-readable identifier for a developer on the Google Play Store \\ \hline
        \raggedright instrument (P1303) & \raggedright drum & musical instrument that a person plays or teaches or used in a music occupation \\ \hline
        \raggedright index case of (P1677) & \raggedright Ebola virus disease in the United States & primary case, patient zero: initial patient in the population of an epidemiological investigation \\ \hline
        \raggedright reward program (P4446) & \raggedright Chase Ultimate Rewards & reward program associated with the item \\ \hline
        \raggedright charge (P1595) & \raggedright seditious libel & offence with which someone is charged, at a trial \\ \hline
        \raggedright had as last meal (P3902) & \raggedright ice cream & components of the last meal had by a person before death \\ \hline
        \raggedright relevant qualification (P4968) & \raggedright Project Management Professional & practitioners of this industry get this degree, licence or certification after specialist education, apprenticeship, or professional review. This includes qualifications that one needs to obtain in order to join the industry or qualifications that one obtains after a certain level of experience in the industry in order to progress further in their career. \\ \hline
        \raggedright for work (P1686) & \raggedright Little Life & qualifier of award received (P166) to specify the work that an award was given to the creator for \\ \hline
        \raggedright interested in (P2650) & \raggedright The Holocaust & item of special or vested interest to this person or organisation \\ \hline
        \raggedright health specialty (P1995) & \raggedright ophthalmology & main specialty that diagnoses, prevent human illness, injury and other physical and mental impairments \\ \hline
        \raggedright appears in the form of (P4675) & \raggedright bull & this fictional or mythical entity takes the form of that entity \\ \hline
        \raggedright vessel class (P289) & \raggedright Concordia-class cruise ship & series of vessels built to the same design of which this vessel is a member \\ \hline
        \raggedright medical condition treated (P2175) & \raggedright non-small-cell lung carcinoma & disease that this pharmaceutical drug, procedure, or therapy is used to treat \\ \hline
        \raggedright sports discipline competed in (P2416) & \raggedright marathon & discipline an athlete competed in within a sport \\ \hline
        \raggedright madhhab (P9929) & \raggedright Shafi`i & Islamic school of thought within Fiqh \\ \hline
        \raggedright cause of death (P509) & \raggedright nitric acid poisoning & underlying or immediate cause of death.  Underlying cause (e.g. car accident, stomach cancer) preferred.  Use 'manner of death' (P1196) for broadest category, e.g. natural causes, accident, homicide, suicide \\
\end{longtblr}

\section{Training parameters of entity recognizer}
\label{parameters}

Table \ref{tbl:Params} details the parameters employed to train the privacy-oriented entity recognition model from Section \ref{sec:er}.

\begin{table}[h]
\caption{\label{tbl:Params} Training Parameters for the RoBERTa model}
\begin{tabular}{p{5cm}p{4cm}}  
	\toprule  
     Optimizer & AdamW  \\ 
     Learning rate & 2e-5\\ 
     Loss function & CrossEntropy \\
     Inference layer & Linear \\
     Epochs & 3 \\
     Full fine-tuning & yes \\
     GPU & yes \\
     Early stopping & yes
\end{tabular}
\end{table}

\vspace{-5mm}
\section{Label Agreement}
\label{label_confusion}

\begin{figure}[h]
\centering
\includegraphics[width = 0.95\textwidth]{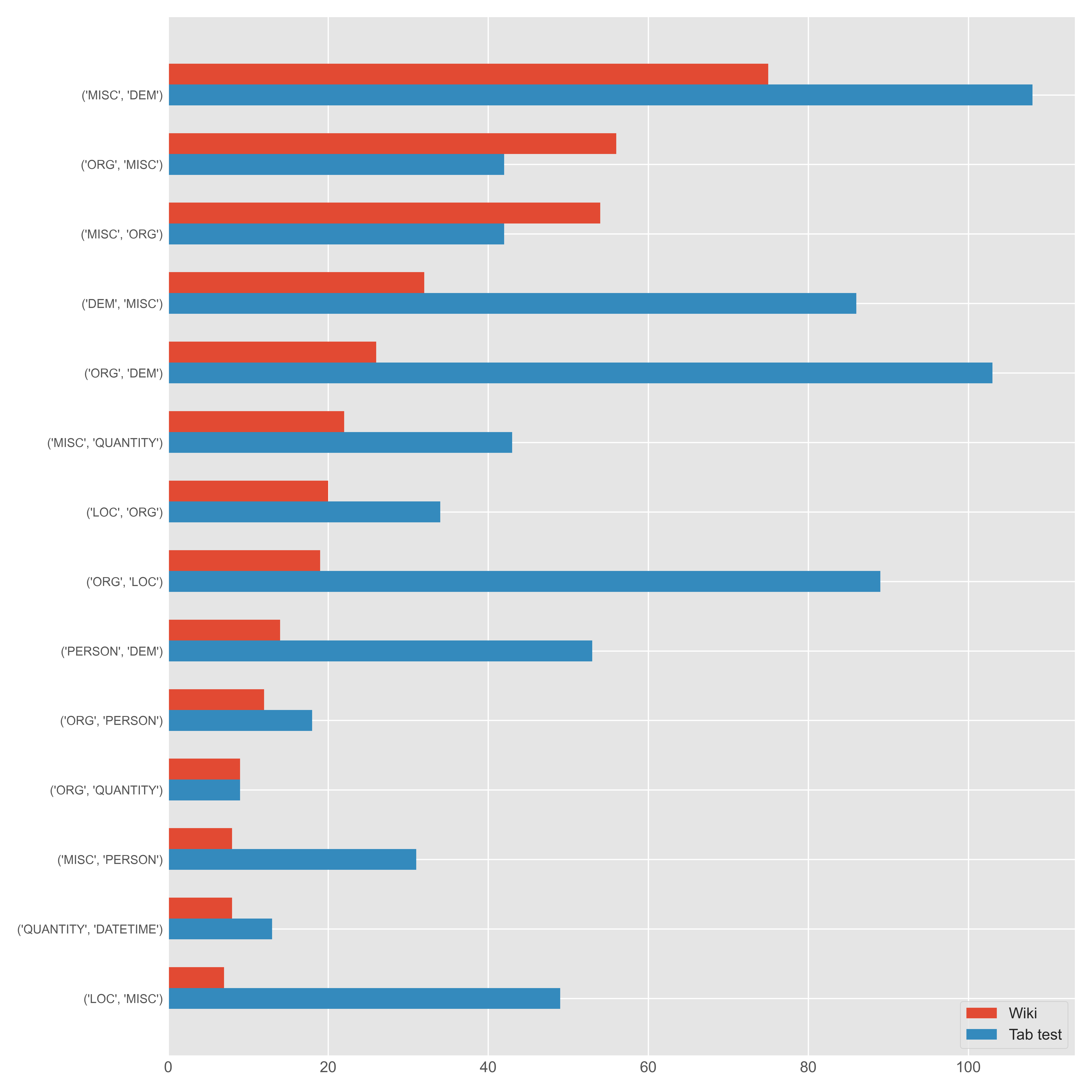}
\caption{Most common label confusion pairs common in the test sets of the annotated Wikipedia biographies and the TAB corpus. The first element of the pair corresponds to the gold standard label and the second to the output from the entity recognizer.}
\label{fig:label_graph}
\end{figure}

Frequently confused label pairs (see Section \ref{ssec:ld}) are shown in Figure \ref{fig:label_graph}.

\section{LLM probabilities: base models}
\label{model}

Table \ref{tbl:tabular_models} describes the (ordered) based models the Autogluon tabular predictor employs for the LLM-probability based approach of Section \ref{ssec:tabular}

\begin{table}[h]
\centering
\caption{\label{tbl:tabular_models} Base models of the Tabular predictor in the order they are trained when using the AutoGluon library. This order based on training time and reliability to ensure efficient training time \citep{autogluon}.}
\begin{tabular}{lp{8cm}}  
	\toprule  
	 \textbf{Model name} & \textbf{Description}\\
	\midrule 
 \textit{KNeighborsUnif} & k-nearest neighbors classifier with uniform weighting\\
    \textit{KNeighborsDist}& k-nearest neighbors classifier with distance weighting (the inverse of their distance)\\
    \textit{LightGBMXT}& LightGMB model that enables extra trees\\
    \textit{LightGBM}& default light gradient-boosting machine implementation\\
     \textit{RandomForestGini}& random forest classifier with Gini impurity loss function\\
     \textit{RandomForestEntr}& random forest classifier with entropy loss function\\
     \textit{CatBoost}& algorithm that performs gradient boosting on decision trees\\
     \textit{ExtraTreesGini}& extra-trees classifier with Gini impurity loss function\\
     \textit{ExtraTreesEntr}& extra-trees classifier with entropy loss function\\
     \textit{NeuralNetFastAI}& fast AI neural network for classification of tabular data\\
     \textit{XGBoost}& an optimized distributed gradient boosting decision tree model \\
     \textit{LightGBMLarge}& LightGMB model that enables larger models\\
     \textit{NeuralNetTorch}& PyTorch neural network for classification of tabular data\\
     \textit{WeightedEnsemble\_L2}& meta-model that implements Ensemble Selection of the above models\\
\end{tabular}
\end{table}

\section{Training size and performance}
\label{size}

\begin{figure}[h]
  \centering
{\includegraphics[width=0.49\textwidth]{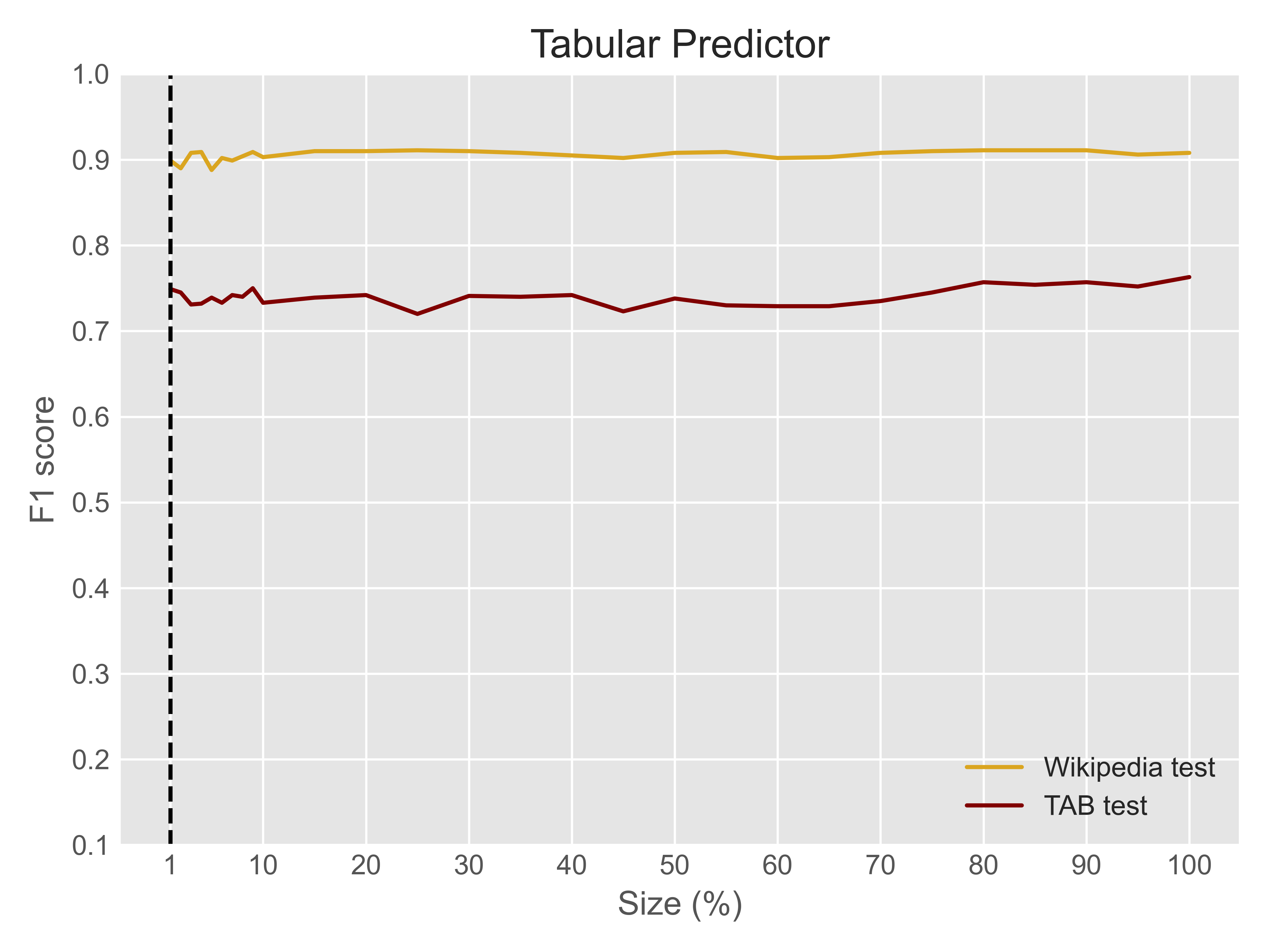}}
  \hfill
{\includegraphics[width=0.49\textwidth]{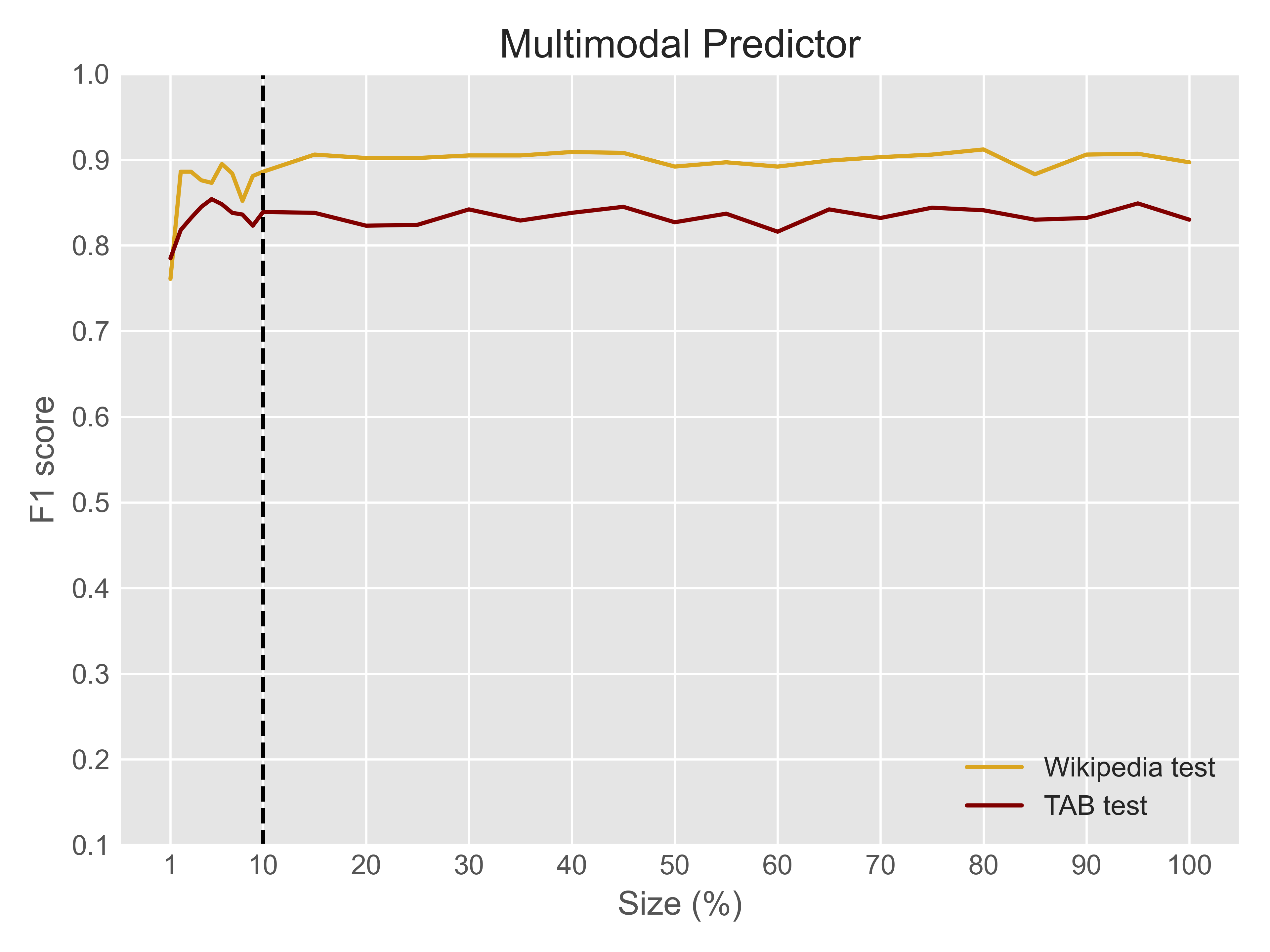}}
  \caption{Performance of the tabular and multimodal predictors when different training sizes are used during training. We report the F1 score for the annotated Wikipedia test dataset and the TAB test dataset as well.}
  \label{fig:training_size}
\end{figure}

Figure \ref{fig:training_size} shows the $F_1$ score of both the Tabular and the Multimodal Autogluon predictors (LLM probabilities Section \ref{ssec:scrm} and span classification Section \ref{ssec:ppa} respectively) at different training sizes for both datasets. We use a random sample of 1\% to 100\% for each training dataset split.

\section{Perturbation thresholds}
\label{per_graphs}

\begin{figure}[h]
  \centering
{\includegraphics[width=0.49\textwidth]{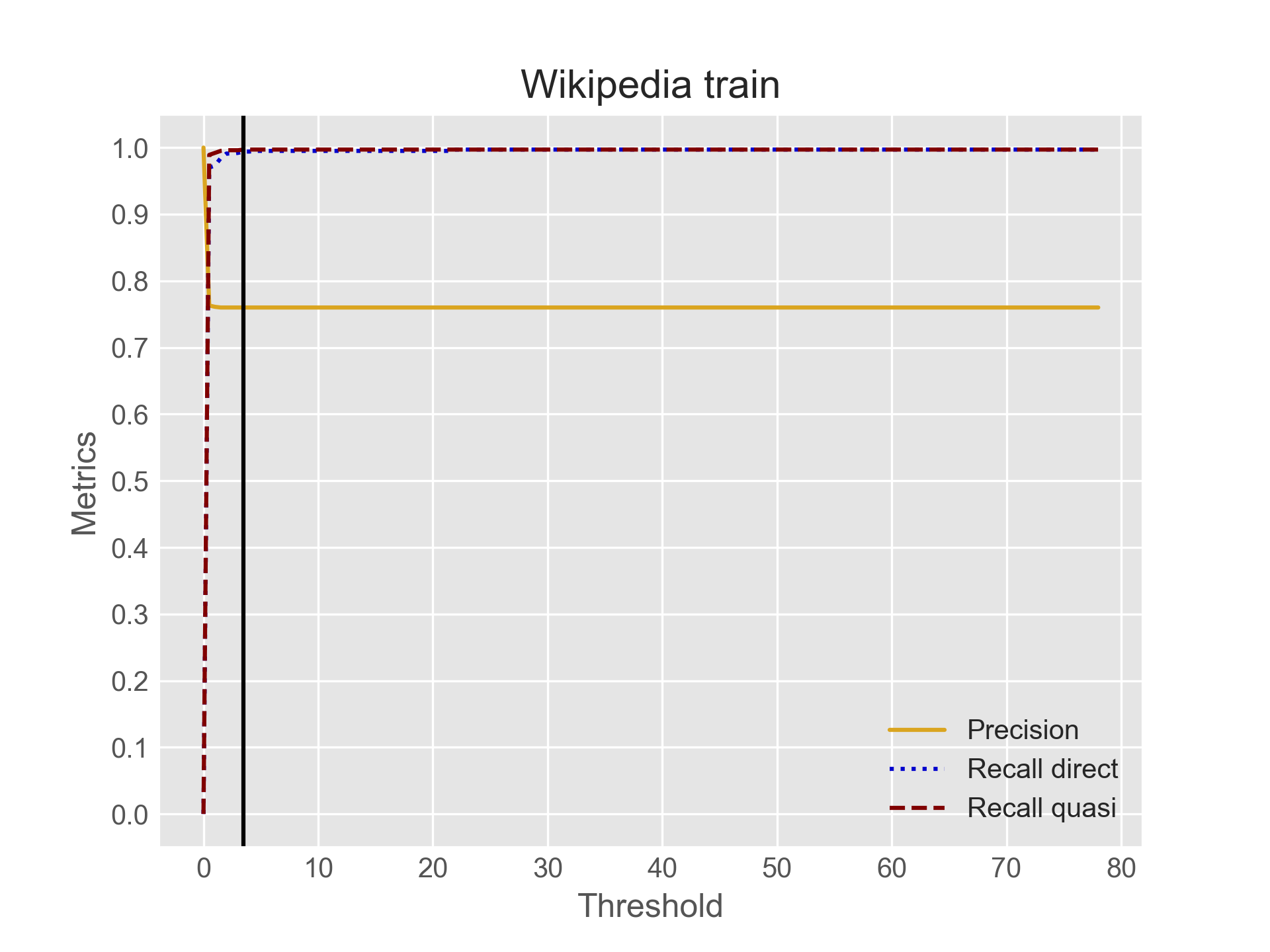}}
  \hfill
{\includegraphics[width=0.49\textwidth]{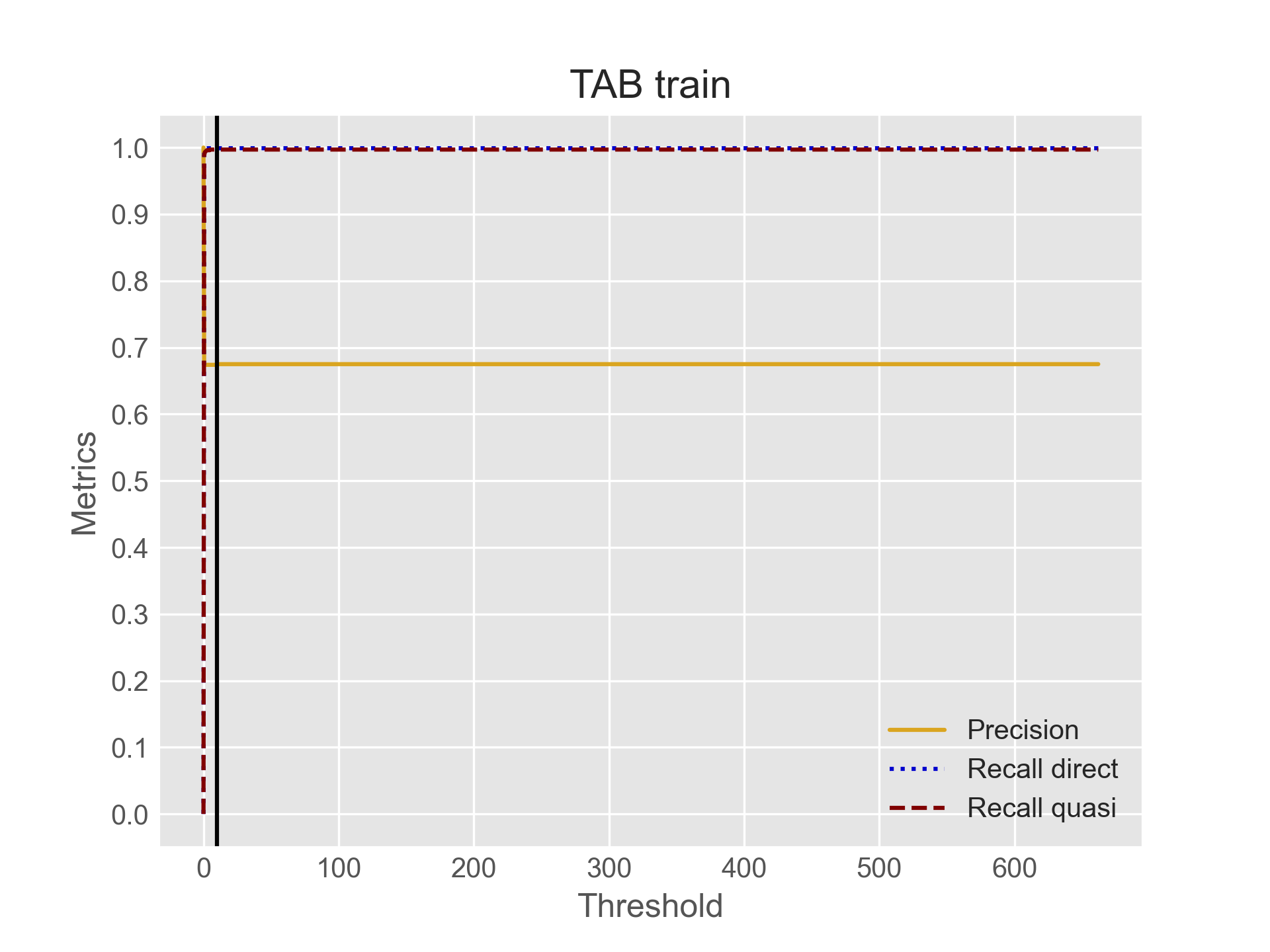}}
  \caption{Precision and recall score for direct and quasi identifiers at different thresholds of probability difference for the Wikipedia and TAB train datasets. The black line indicates the threshold where the cost function is maximized. This is approximately 3.5 for Wikipedia and 10 for TAB)}
  \label{fig:perturbation_graph}
\end{figure}

Figure \ref{fig:perturbation_graph} shows the performance of different perturbation thresholds for both datasets for the training dataset split, with the black line indicating the threshold used in Section \ref{ssec:petrubation} for evaluation.

\clearpage
\bibliography{sn-bibliography}

\end{document}